\documentclass{article} 
\usepackage{ms_tech_report}
\usepackage[colorlinks = true,
            linkcolor = blue,
            urlcolor  = blue,
            citecolor = blue,
            anchorcolor = blue]{hyperref}   
\usepackage{microtype}
\usepackage{hyperref}
\usepackage{url}
\usepackage{booktabs}
\usepackage{enumitem}
\usepackage{multicol}
\usepackage{multirow}
\usepackage{CJKutf8}
\usepackage{amsmath}
\usepackage{siunitx}
\usepackage{floatflt}
\usepackage{wrapfig}
\usepackage{fontawesome5}
\usepackage[most]{tcolorbox}
\usepackage{xcolor}
\definecolor{derivecolor}{RGB}{70, 90, 130}
\usepackage{seqsplit}
\newcommand{\filepath}[1]{\texttt{\small\seqsplit{#1}}}
\DeclareRobustCommand{\captionfilepath}[1]{%
  {\small\ttfamily\seqsplit{#1}}%
}
\usepackage{booktabs}
\usepackage{caption}
\usepackage{subfigure}

\newcommand{\faHuggingFace}{\raisebox{-0.2\height}{\includegraphics[height=0.9em]{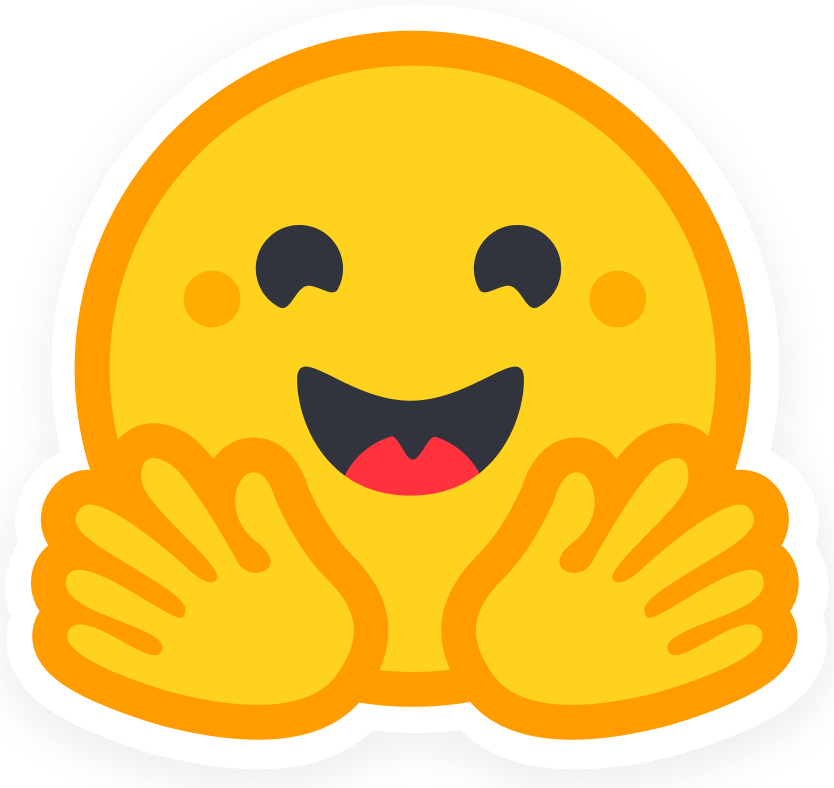}}}

\colmfinalcopy

\newcommand{\corrauth}{\textsuperscript{\faEnvelope[regular]}}

\title{Synthetic Computers at Scale for Long-Horizon Productivity Simulation}

\author{Tao Ge\,\corrauth, Baolin Peng\,\corrauth, Hao Cheng, Jianfeng Gao \\
Microsoft \\ 
\faEnvelope[regular]~ \texttt{\{taoge, baolinpeng\}@microsoft.com} \\
\faHuggingFace~ \href{https://huggingface.co/datasets/microsoft/synthetic-computers-at-scale}{huggingface.co/datasets/microsoft/synthetic-computers-at-scale}
}

\begin{document}
\maketitle
\vspace{-0.2cm}
\begin{abstract}



Realistic long-horizon productivity work is strongly conditioned on user-specific computer environments, where much of the work context is stored and organized through directory structures and content-rich artifacts. To scale synthetic data creation for such productivity scenarios, we introduce Synthetic Computers at Scale, a scalable methodology for creating such environments with realistic folder hierarchies and content-rich artifacts (e.g., documents, spreadsheets, and presentations). Conditioned on each synthetic computer, we run long-horizon simulations: one agent creates productivity objectives that are specific to the computer's user and require multiple professional deliverables and about a month of human work; another agent then acts as that user and keeps working across the computer---for example, navigating the filesystem for grounding, coordinating with simulated collaborators, and producing professional artifacts---until these objectives are completed.

\vspace{0.2cm}

In preliminary experiments, we create 1,000 synthetic computers and run long-horizon simulations on them; each run requires over 8 hours of agent runtime and spans more than 2,000 turns on average. These simulations produce rich experiential learning signals, whose effectiveness is validated by significant improvements in agent performance on both in-domain and out-of-domain productivity evaluations. Given that personas are abundant at billion scale, this methodology can in principle scale to millions or even billions of synthetic user worlds with sufficient compute, enabling broader coverage of diverse professions, roles, contexts, environments, and productivity needs. We argue that scalable synthetic computer creation, together with at-scale simulations, is highly promising as a foundational substrate for agent self-improvement and agentic reinforcement learning in long-horizon productivity scenarios.
\end{abstract}

\begin{figure}[!h]
    \centering
    \includegraphics[width=14.5cm, height=6.5cm]{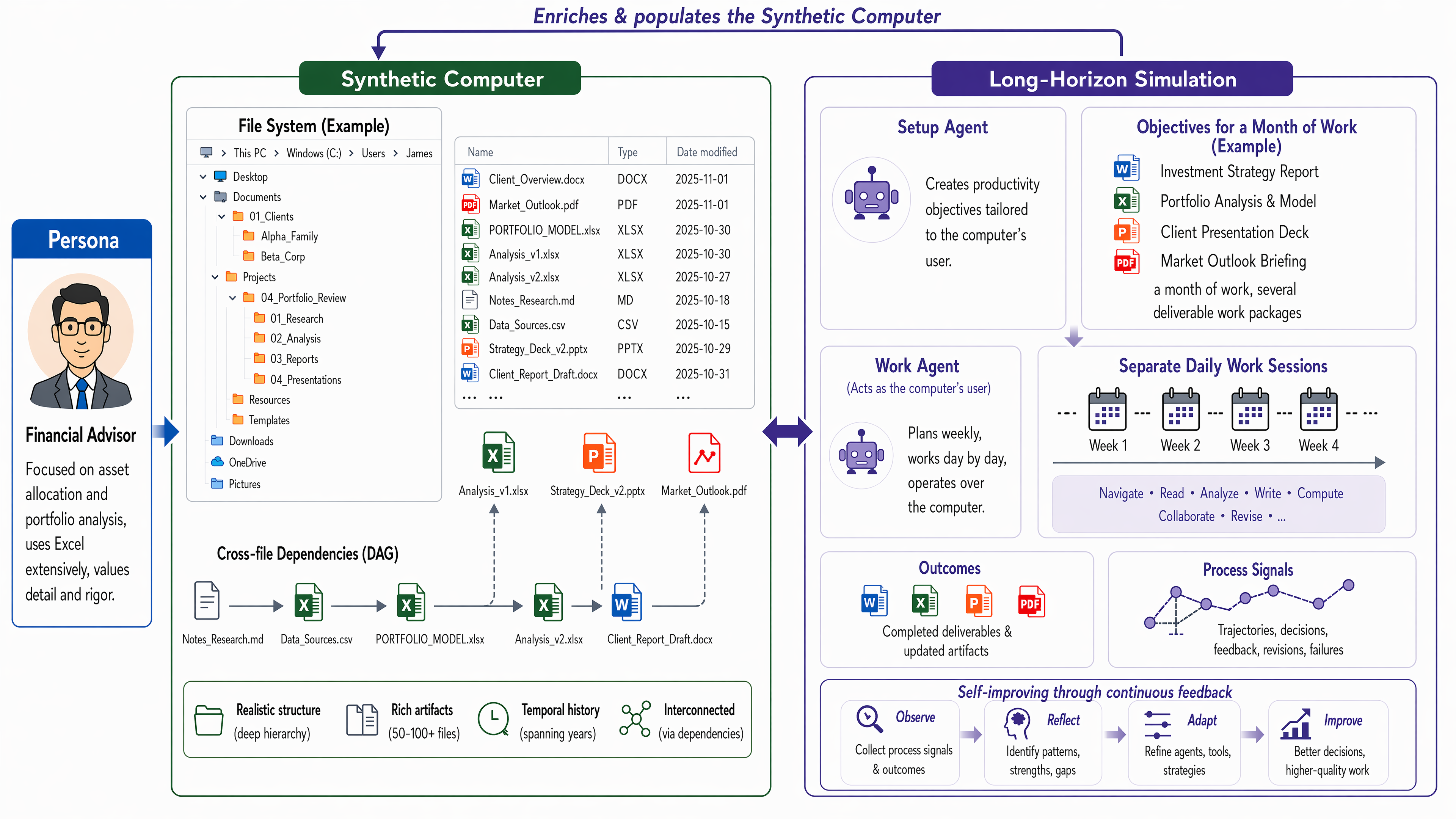} \vspace{-0.2cm}
    \caption{Overview of our methodology: We create user-specific synthetic computers from personas and use them as grounding environments for long-horizon productivity simulations, producing both professional deliverables and process signals for improving agents in productivity scenarios.}    \label{fig:overview}
\end{figure}

\section{Introduction}

As AI agents expand from conversation-bounded assistants (e.g., \href{https://chat.com/}{ChatGPT}) to repository-grounded coding agents (e.g., \href{https://cursor.com/}{Cursor}) and now toward long-horizon agents grounded in entire user computers (e.g., \href{https://claude.com/product/cowork}{Claude Cowork}), they are increasingly expected to support realistic productivity work: gathering information, analyzing evidence, coordinating with others, and producing artifacts\footnote{In this work, we use ``artifacts'' primarily to refer to structured productivity files, such as documents, spreadsheets, presentations, and PDFs, rather than plain-text-only files.} such as documents, spreadsheets, presentations, reports, and supporting materials. To perform this work well, agents need increasingly rich, user-specific context to stay grounded~\citep{openai2026gpt55,anthropic2026opus47,jimenez2023swe,patwardhan2025gdpval,maharana2024evaluating}. When agents can act over this context for long horizons, the resulting long-horizon trajectories provide rich experiential data for improving both run-time agent behavior and the capabilities of the underlying LLMs in realistic productivity scenarios~\citep{cheng2026lifebenchbenchmarklonghorizonmultisource,li2026horizonbenchlonghorizonpersonalizationevolving,laban2026llms}. In practice, however, real trajectories are costly to collect at scale because they are usually grounded in private computer environments containing personal artifacts, enterprise documents, project state, and interaction history~\citep{ozaki2025privacy,jian2026cua}. This makes synthetic data necessary, but difficult to create well: realistic user environments are complex, heterogeneous, and artifact-rich, with diverse directory structures, professional artifacts, and accumulated work history~\citep{feng2025webworldmodels,copet2025cwm,cai2025autoforgeautomatedenvironmentsynthesis}.

Together, these points lead to three guiding principles that motivate this work:
\begin{itemize}[leftmargin=*]
    \item \textbf{Productivity work is context-heavy by nature.} Realistic productivity work is grounded in existing files, project history, prior decisions, collaborator feedback, and evolving work state.
    \item \textbf{The key challenge is using rich user context over long horizons.} For productivity agents, success depends not only on solving isolated tasks, but on using the user's files, history, and evolving work context effectively over long horizons.
    \item \textbf{Synthetic data must synthesize the context, not only the task.} Without realistic user environments to condition on, synthetic data degenerates into generic, toy workflows that remain far from real work scenarios.
\end{itemize}



To scale synthetic data creation for realistic long-horizon productivity scenarios, we introduce Synthetic Computers at Scale, a methodology for creating diverse, artifact-rich, user-specific synthetic computer environments. Building on our prior persona-driven synthetic data creation methodology~\citep{ge2024scaling}, we start from large pools of personas and use large language models (LLMs) to progressively elaborate each sampled persona into a user-specific computer environment populated with realistic directory structures and content-rich artifacts, as shown in Figure \ref{fig:overview}.

With each synthetic computer in place, we use it to run a long-horizon productivity simulation. Conditioned on the user profile and the computer's contents, a setup agent creates productivity objectives tailored to the computer's user. These objectives typically require multiple challenging professional deliverables, such as reports, spreadsheets, and presentations, and correspond to about a month of human work. A separate work agent then acts as that user in the simulation, operating over the computer to complete the objectives: navigating the filesystem for grounding, using relevant artifacts and references, coordinating with simulated collaborators, incorporating feedback, and iteratively creating or revising the required deliverables.

In our preliminary experiments, we instantiate 1,000 synthetic computers and run one long-horizon simulation per computer. Each simulation corresponds to about a month of work for the computer’s user, requires over 8 hours of agent runtime, and spans more than 2,000 turns on average. These simulations produce rich experiential learning signals from both process and outcome: intermediate trajectories record how agents search, plan, revise, coordinate with collaborators, incorporate feedback, and recover from failures, while final deliverables provide outcome-level signals on the quality of the completed work. We validate the effectiveness of these signals through both in-domain and out-of-domain evaluations, demonstrating significant improvements in agent performance on long-horizon productivity tasks.

Given that personas are abundant at billion scale~\citep{ge2024scaling}, this methodology can in principle scale to millions or even billions of synthetic computers with sufficient compute, enabling broader coverage of diverse professions, roles, contexts, environments, and productivity needs. We argue that scalable synthetic computer creation, together with at-scale simulations, provides a promising foundational substrate for agent self-improvement and agentic reinforcement learning in long-horizon productivity scenarios.

To support research on this methodology, we release 100 synthetic computers---50 Windows-style and 50 macOS-style environments---as well as retrospective analysis reports for 500 long-horizon simulations. These resources are intended to support further study of synthetic computer creation, experiential learning from long-horizon trajectories, and scalable productivity simulation.

\section{Synthetic Computer Creation}\label{sec:computer}



\begin{figure}[htbp]
    \centering
    \includegraphics[width=15cm, height=3.2cm]{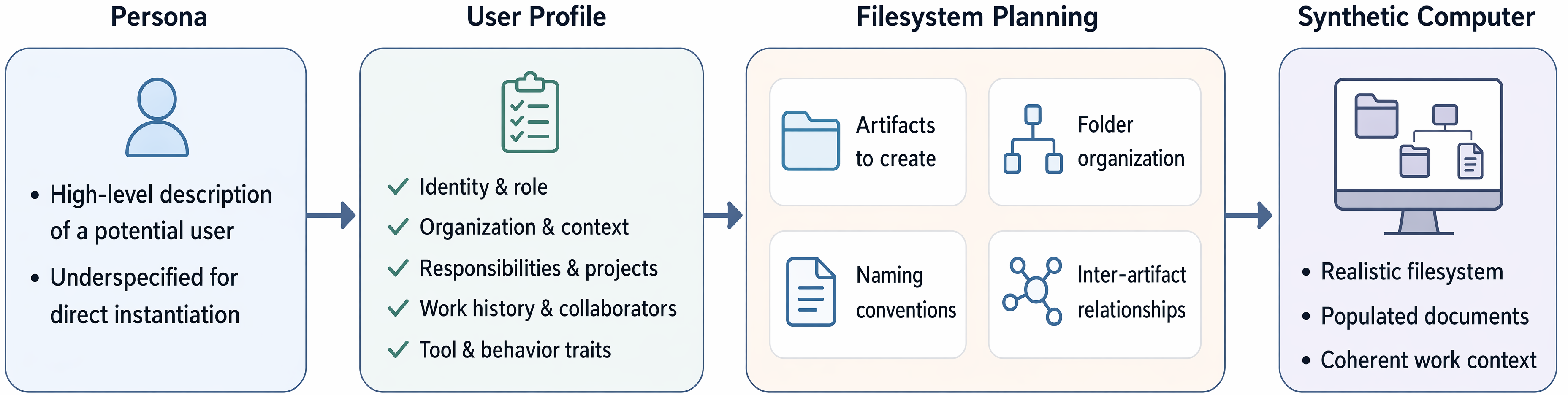}
    \caption{Overview of synthetic computer creation. A persona is first expanded into a detailed user profile, which is then used to plan the filesystem, artifacts, folder organization, naming conventions, and inter-artifact relationships before instantiating a populated synthetic computer.}    \label{fig:sec2_overview}
\end{figure}

We create each synthetic computer by progressively elaborating a sampled persona into a populated computer environment, as shown in Figure \ref{fig:sec2_overview}: the persona is first expanded into a detailed user profile (Section~\ref{subsec:profile}); this profile is used to plan the filesystem of the computer environment (Section~\ref{subsec:planning}); and the planned filesystem is then instantiated as a directory hierarchy populated with realistic, content-rich artifacts (Section~\ref{subsec:artifact}).

\subsection{Persona-Driven User Profiles}
\label{subsec:profile}

\begin{tcolorbox}[
    colback=blue!5,           
    colframe=blue!50!black,   
    arc=3mm,                  
    boxrule=0.5pt,            
    left=2mm, right=2mm, top=1mm, bottom=1mm,
    title=\textbf{Persona},
    fontupper=\small,         
    fonttitle=\small\bfseries
]
A financial advisor focused on assessing how different asset classes are likely to perform over time, and on applying that insight to make well-informed investment choices. They are probably acquainted with the Vanguard Capital Markets Model\textsuperscript{\textregistered} (VCMM) and its forecasting capabilities, and want to use those projections to guide how they balance their portfolio between shares and bonds. They are also interested in Vanguard's recommended asset allocations and in evaluating how those allocations fit their particular investment goals and risk tolerance.
\end{tcolorbox}

Each synthetic computer starts from a persona. As shown in the example above, a persona provides high-level information about a potential user, but by itself is too underspecified to determine what should exist on that user's computer. We therefore first expand the persona into a detailed user profile that serves as the semantic foundation for the synthetic computer:

\begin{tcolorbox}[
    colback=blue!5,           
    colframe=blue!50!black,   
    arc=2mm,
    boxrule=0.4pt,
    left=3mm, right=3mm, top=1mm, bottom=1mm,
    title=\textbf{User Profile},
    fonttitle=\small\bfseries,
    fontupper=\small,         
    breakable
]
\textbf{Identity.} Margaret Elaine Forsythe (\texttt{mforsythe}), based in Denver, Colorado.

\smallskip
\textbf{Occupation.} Senior Financial Advisor.

\smallskip
\textbf{Organization.} Meridian Wealth Partners.

\smallskip
\textbf{Career stage.} Senior; 16+ years in portfolio management and long-term investment planning, specializing in capital market forecasting and asset allocation strategy. CFA charterholder (2013); CFP certified (2010).

\smallskip
\textbf{Responsibilities.} Constructing and monitoring client portfolios across equity and fixed income; interpreting VCMM 10-year return projections into allocation recommendations; conducting Investment Policy Statement reviews; presenting strategy in quarterly client meetings; leading firm investment committee discussions on capital market assumptions; mentoring junior advisors.

\smallskip
\textbf{Recent work history.} \emph{2024 H1:} led firm-wide rebalancing initiative using updated VCMM projections after the Fed rate plateau. \emph{2024 H2:} developed structured client risk reassessment process incorporating Monte Carlo and VCMM outputs across 40+ households. \emph{2025 H1:} refreshed firm strategic asset allocation framework and evaluated alternatives integration. \emph{2025 H2:} rewrote IPS templates for 28 client accounts; began ESG overlay research. \emph{2026 Q1:} presented updated 10-year capital market assumptions to investment committee.

\smallskip
\textbf{Current projects.} (i) 2026 VCMM-driven model portfolio refresh; (ii) onboarding of three new HNW clients ($\sim$\$18M AUM); (iii) systematic rebalancing trigger framework; (iv) alternatives integration research (REITs, commodities, liquid alts); (v) ESG equity overlay evaluation.

\smallskip
\textbf{Collaborators.} David Hartley (manager, Managing Director); Kevin Tran (direct report, junior associate); Sandra Okonkwo (peer, fixed-income specialist); James Whitfield (compliance officer); Patricia Huang (external, Vanguard advisor services); ...

\smallskip
\textbf{Common work products.} Investment Policy Statements; capital markets outlook memos; client portfolio review decks (PowerPoint); asset allocation analysis spreadsheets (Excel); Monte Carlo simulation summaries; quarterly performance reports; risk tolerance questionnaires; internal research memos and white papers.

\smallskip
\textbf{Technical level.} Intermediate.

\smallskip
\textbf{Computer usage level.} High.

\smallskip
\textbf{Preferred tools.} Excel for analytical modeling; Word for IPS drafts and research memos; PowerPoint for client decks; PDF for client distribution. Inputs sourced from Vanguard's advisor portal (VCMM PDFs), Bloomberg terminal exports, and Morningstar Direct.

\smallskip
\textbf{Document habits.} Drafts analysis in Excel before summarizing in Word or PowerPoint. Reviews own Word documents multiple times before sharing; uses Track Changes for collaborative edits. Prints final client-facing documents to PDF before distribution. Tends to over-document.

\smallskip
\textbf{Spreadsheet usage.} Heavy. Builds scenario-analysis tabs and data tables; adds cell comments to document assumptions and data sources; saves date-stamped intermediate versions before major structural changes.

\smallskip
\textbf{Attachment-saving behavior.} Retains source PDFs (VCMM reports, Vanguard white papers, client account statements) and vendor data exports (Bloomberg CSVs, Morningstar reports) as inputs to her Excel models.

\smallskip
\textbf{Naming preferences.} Descriptive filenames; uses explicit version suffixes; rarely keeps default names.

\smallskip
\textbf{Organization style.} Tidy and systematic, but occasionally accumulates multiple versions of the same spreadsheet without consistent archiving discipline.
\end{tcolorbox}

The user profile captures both professional context and computer-use behavior. It includes the user's identity, occupation, organization, career stage, responsibilities, recent work history, current projects, collaborators, and common work products. It also records traits that are directly relevant to filesystem construction, such as technical level, computer usage level, tidiness, preferred tools, document habits, spreadsheet usage, attachment-saving behavior, naming preferences, and organization style. These details help determine not only what artifacts should appear on the computer, but also where they are likely to be stored, how they are named, and how they relate to the user's ongoing work.


\subsection{Computer Environment Planning}
\label{subsec:planning}

Given the user profile, we next plan the filesystem of the user-specific computer environment. The resulting plan specifies the directory organization, virtual time axis, project structure, file inventory, artifact types, and cross-file dependencies before any file contents are generated.

\subsubsection{Filesystem Policy Generation}
\label{subsubsec:filesystem_policy}

The first step in computer environment planning is to generate a user-specific filesystem policy from the user profile. This policy specifies the basic conventions under which the synthetic computer is organized, including the system start time, drive layout, default paths, storage patterns, organization style, naming style, and usage patterns:

\begin{tcolorbox}[
    colback=blue!5,           
    colframe=blue!50!black,   
    arc=2mm,
    boxrule=0.4pt,
    left=3mm, right=3mm, top=2mm, bottom=2mm,
    title=\textbf{Filesystem Policy},
    fonttitle=\small\bfseries,
    fontupper=\small,
    breakable
]
\textbf{System start.} 2022-11-05 17:41.

\smallskip
\textbf{Drive layout.} \texttt{C:} (system); \texttt{D:} (data).

\smallskip
\textbf{Default user paths.} Desktop, Documents, Downloads, Pictures, and AppData under \texttt{C:/Users/mforsythe/}.

\smallskip
\textbf{Storage patterns.} Office files and project work go to \texttt{D:/ClientWork}, \texttt{D:/Research}, and \texttt{D:/ModelPortfolios}. Downloads land in \texttt{C:/Users/mforsythe/Downloads} or \texttt{D:/Research/ExternalData}. Screenshots go to \texttt{C:/Users/mforsythe/Pictures/Screenshots}. Temporary files often end up in Downloads or on the Desktop.

\smallskip
\textbf{Organization style.} High tidiness; uses project folders; low desktop and download clutter; does not pile files into default folders.

\smallskip
\textbf{Naming style.} Consistent and descriptive; uses version suffixes; rarely keeps default filenames. Representative examples: \texttt{IPS\_Draft\_v1.docx}, \texttt{AllocationModel\_v2.xlsx}, \texttt{ClientReview\_Q1\_2025.pptx}, \texttt{CapMarketsOutlook\_2026\_FINAL.pdf}.

\smallskip
\textbf{Usage patterns.} High computer usage; frequently downloads files, creates new documents, edits existing files, and switches between projects; rarely takes screenshots.
\end{tcolorbox}

\subsubsection{Filesystem Planning}
\label{subsubsec:filesystem_planning}

Given the user profile and filesystem policy, we next plan the files that should appear on the synthetic computer. This step uses the user's current projects, responsibilities, collaborators, and document habits to infer a coherent file inventory, including logical paths, artifact types, descriptions, timestamps, origins, and content modes. The resulting plan specifies not only which files exist, but also how they are organized within the directory structure (see below) and how they relate to the user's accumulated work context:

\begin{tcolorbox}[
    colback=blue!5,           
    colframe=blue!50!black,   
    arc=2mm,
    boxrule=0.4pt,
    left=3mm, right=3mm, top=2mm, bottom=2mm,
    title=\textbf{Directory Tree},
    fonttitle=\small\bfseries,
    fontupper=\footnotesize,
    breakable
]
\begin{minipage}[t]{0.48\linewidth}
\begin{verbatim}
C:/
+-- Users/
    +-- mforsythe/
        |-- AppData/
        |   +-- Roaming/
        |       +-- Microsoft/
        |           +-- Excel/
        |               +-- XLSTART/
        |-- Desktop/
        |-- Documents/
        |   |-- Admin/
        |   |-- FirmPolicies/
        |   |-- PersonalFinance/
        |   |-- Templates/
        |   +-- WhitePaper/
        +-- Downloads/
\end{verbatim}
\end{minipage}%
\hfill
\begin{minipage}[t]{0.48\linewidth}
\begin{verbatim}
D:/
|-- ClientWork/
|   |-- Castellano_Robert/
|   |-- ExistingClients/
|   |-- FirmReference/
|   |-- InvestmentCommittee/
|   |-- InvestmentPolicies/
|   +-- RiskProfiles/
|-- ModelPortfolios/
|   |-- Archive/
|   |-- Exemplars/
|   |-- RebalancingFramework/
|   +-- VCMM_2026/
+-- Research/
    |-- Alternatives/
    |-- ESG/
    |-- ExternalData/
    +-- VCMM/
\end{verbatim}
\end{minipage}
\end{tcolorbox}

In addition to the file inventory, we construct a directed dependency graph over planned files. The graph captures relations such as one file referencing another, being derived from another, representing a later version, or being extracted from an archive. A central purpose of this dependency graph is to avoid treating files as independent samples conditioned only on the user profile. Real computer environments contain correlated artifacts: later files often reuse, summarize, revise, or cite earlier ones. The dependency graph makes these relationships explicit and provides the conditioning structure for artifact instantiation, so that each file can be generated with reference to the earlier materials it builds on. For example, in the financial-advisor computer, an internal VCMM asset-class projection workbook (\texttt{D:/Research/VCMM/VCMM\_AssetClassProjections\_2025.xlsx}) may be derived from a downloaded Vanguard return-projections PDF, and later allocation-model spreadsheets can build on that workbook:

\begin{tcolorbox}[
    colback=blue!5,           
    colframe=blue!50!black,   
    arc=2mm,
    boxrule=0.4pt,
    left=3mm, right=3mm, top=2mm, bottom=2mm,
    title=\textbf{File List (Excerpt)},
    fonttitle=\small\bfseries,
    fontupper=\footnotesize,
    breakable
]
\setlength{\parskip}{4pt}
\setlength{\parindent}{0pt}
\newcommand{\fileentry}[2]{%
  \texttt{\small #1}\\
  \hspace*{1.5em}\begin{minipage}[t]{\dimexpr\linewidth-1.5em\relax}#2\end{minipage}\par
}
\fileentry{C:/Users/mforsythe/AppData/Roaming/Microsoft/Excel/XLSTART/PersonalMacros.xlsm}{%
\emph{2023-05-10} \quad Excel personal macro workbook with VBA macros for formatting allocation tables and chart styles. Auto-loads with Excel on startup.}
\fileentry{C:/Users/mforsythe/Documents/WhitePaper/VCMM\_ClientAllocationFramework\_v1.docx}{%
\emph{2024-11-15} \quad First draft of internal white paper on integrating VCMM projections into client allocation frameworks. $\sim$22 pages.}
\fileentry{D:/Research/VCMM/VCMM\_ReturnProjections\_Summary\_2025.pdf \normalfont\quad\emph{(web download)}}{%
\emph{2025-01-10} \quad Vanguard Capital Markets Model 2025 return projections summary, downloaded from Vanguard's advisor portal.}
\fileentry{D:/Research/VCMM/VCMM\_AssetClassProjections\_2025.xlsx}{%
\emph{2025-01-15} \quad Structured workbook compiling VCMM 10-year return projections, transcribed from the Vanguard PDF above. Foundation data for all VCMM-based allocation work.\\
\textcolor{derivecolor}{\textit{$\hookrightarrow$ derived from} \texttt{\small D:/Research/VCMM/VCMM\_ReturnProjections\_Summary\_2025.pdf}}}
\fileentry{D:/ModelPortfolios/VCMM\_2026/AllocationModel\_Conservative\_v1.xlsx}{%
\emph{2025-02-14} \quad First-draft allocation model for the conservative tier, with asset class weights, expected return inputs, and a scenario comparison tab.}
\fileentry{D:/ModelPortfolios/VCMM\_2026/AllocationModel\_Conservative\_v2.xlsx}{%
\emph{2025-07-22} \quad Revised conservative allocation model with mid-year VCMM projections, Monte Carlo summary, and Bloomberg rate data pulls.\\
\textcolor{derivecolor}{\textit{$\hookrightarrow$ derived from} \texttt{\small D:/ModelPortfolios/VCMM\_2026/AllocationModel\_Conservative\_v1.xlsx}}}
\begin{center}
$\vdots$ \quad \emph{(8 more files in \texttt{\small D:/ModelPortfolios/})} \quad $\vdots$
\end{center}
\fileentry{D:/ModelPortfolios/VCMM\_2026/ScenarioAnalysis\_EquityBondSplits\_2025.xlsx}{%
\emph{2025-09-15} \quad Scenario analysis comparing seven equity/bond allocation variants (20/80 through 80/20) using VCMM projections and Bloomberg rate data.\\
\textcolor{derivecolor}{\textit{$\hookrightarrow$ derived from} \texttt{\small D:/Research/VCMM/VCMM\_AssetClassProjections\_2025.xlsx}, \texttt{\small D:/Research/ExternalData/Bloomberg\_RateData\_Q3\_2025.xlsx}}}
\fileentry{D:/ModelPortfolios/VCMM\_2026/CapMarketsOutlook\_2026\_DRAFT.docx}{%
\emph{2025-11-10} \quad Draft 2026 capital markets outlook covering 10-year return projections, macro backdrop, and allocation implications. $\sim$18 pages.\\
\textcolor{derivecolor}{\textit{$\hookrightarrow$ derived from} \texttt{\small D:/ModelPortfolios/VCMM\_2026/ScenarioAnalysis\_EquityBondSplits\_2025.xlsx}, \texttt{\small D:/Research/VCMM/VCMM\_AssetClassProjections\_2025.xlsx}}}
\fileentry{D:/ModelPortfolios/VCMM\_2026/CapMarketsOutlook\_2026\_FINAL.pdf}{%
\emph{2025-12-05} \quad Final PDF of the 2026 outlook, exported after investment committee review. Definitive reference for Meridian's 2026 investment posture.\\
\textcolor{derivecolor}{\textit{$\hookrightarrow$ derived from} \texttt{\small D:/ModelPortfolios/VCMM\_2026/CapMarketsOutlook\_2026\_DRAFT.docx}}}
\begin{center}
\emph{(roughly 50 more files across \texttt{\small D:/ClientWork/}, \texttt{\small D:/Research/ESG/}, \texttt{\small C:/Users/mforsythe/Downloads/}, \ldots)}
\end{center}
\end{tcolorbox}

\subsection{Artifact Creation}
\label{subsec:artifact}

Given the filesystem plan, we instantiate the synthetic computer in two steps. First, we materialize the planned directory structure. Logical paths in the filesystem plan are mapped into a portable on-disk representation while preserving the intended operating-system semantics. For example, a Windows-style path such as \texttt{D:/Research/VCMM/VCMM\_ReturnProjections\_Summary\_2025.pdf} is stored under the corresponding physical directory \texttt{drives/D/Research/VCMM/VCMM\_ReturnProjections\_Summary\_2025.pdf}. We then create all parent directories required by the planned file inventory.

Second, we populate the planned files with content-rich artifacts. Files are instantiated according to their metadata, including their path, description, artifact type, timestamp, origin, and content mode. To respect cross-file dependencies, we use a dependency-aware order~\citep{kahn1962topological} over the file graph: files with no predecessors are instantiated first, and each instantiated file then provides context for its downstream dependents; ties are broken by timestamp to remain consistent with the computer's virtual history.

\begin{figure}
    \centering
    \includegraphics[width=0.98\linewidth]{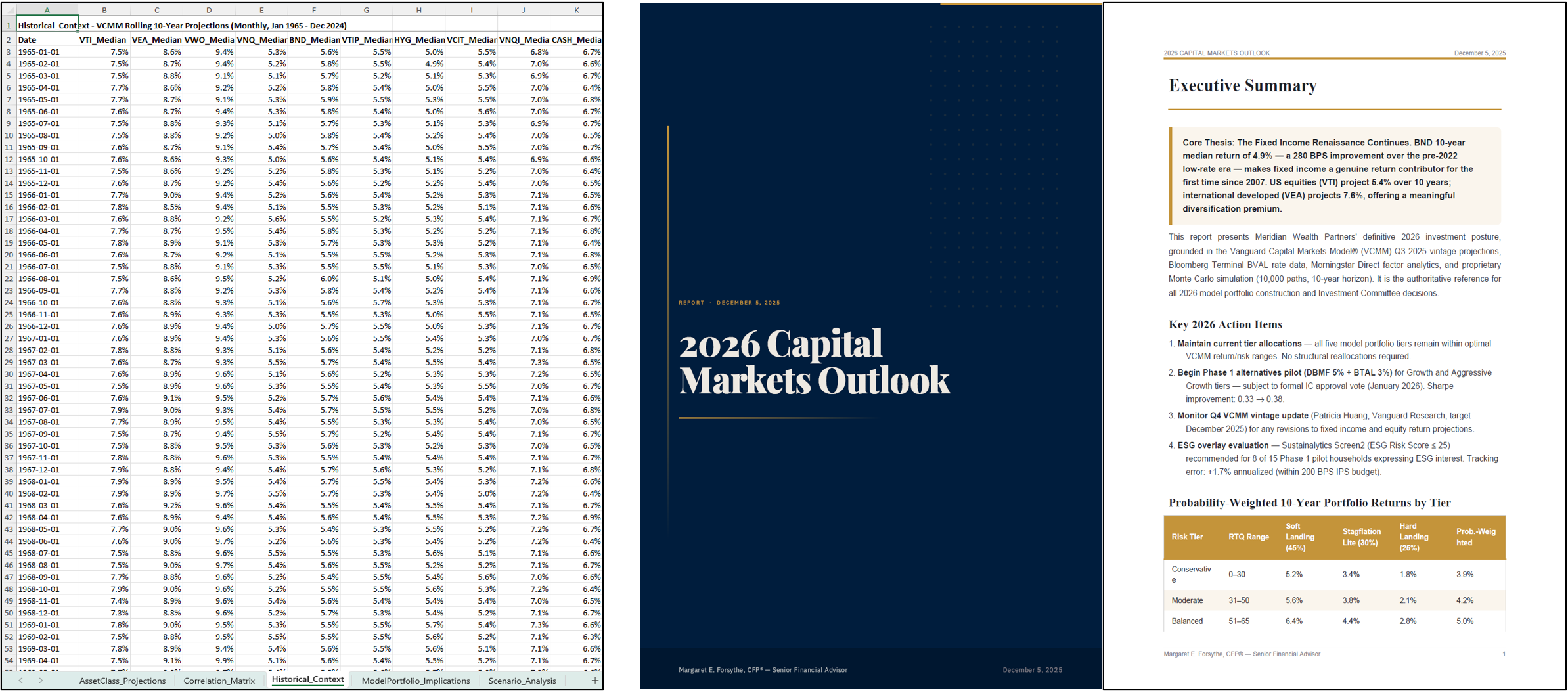}
    \caption{Screenshots of the artifacts created in the synthetic computer. \textbf{Left:} \captionfilepath{D:/Research/VCMM/VCMM\_AssetClassProjections\_2025.xlsx}; \textbf{Right: }\captionfilepath{D:/ModelPortfolios/VCMM\_2026/CapMarketsOutlook\_2026\_FINAL.pdf}}.
    \label{fig:artifact}
\end{figure}

When a file is marked in the plan as a public, web-downloadable artifact, such as \texttt{D:/Research/VCMM/VCMM\_ReturnProjections\_Summary\_2025.pdf}, we first retrieve it from the web and store it at the planned location, falling back\footnote{A more robust planning stage would verify that a candidate public file is actually available and downloadable before marking it as web-downloadable. In the current pipeline, this check is performed after planning, so some planned web downloads fall back to synthesis.} to synthesis only if retrieval fails. Other user-specific artifacts are created by an LLM agent equipped with artifact-creation tools or skills, conditioned on the file description and relevant predecessor artifacts, as shown in Figure \ref{fig:artifact}.

\section{Long-Horizon Productivity Simulation}
\label{sec:simulation}

Given a synthetic computer, we run a long-horizon simulation using two agents with different roles. The first agent, which we call the setup agent, creates productivity objectives tailored to the computer's user and sets up the collaboration setting around those goals (Section~\ref{subsec:setup}). The second agent, which we call the work agent, then acts as that user and carries out the goals by operating over the synthetic computer (Section~\ref{subsec:planning_daily_work_simulation}).

\subsection{Setup}
\label{subsec:setup}

\subsubsection{Productivity Objectives}
\label{subsubsec:productivity_objectives}

The setup agent first determines what productivity objectives should be simulated for the computer's user. Rather than sampling a generic task, the agent conditions on both the user profile and the current state of the computer, including the user's role, responsibilities, active projects, file inventory, and existing artifacts. From this context, it infers\footnote{In practice, we prompt the setup agent to reason from the perspective of someone who understands the user's professional responsibilities, such as a manager, client, or domain expert.} a set of productivity objectives that are realistic and valuable for that user to pursue over about a month of work. These objectives are strongly conditioned on who the user is and what already exists on the computer. They should be challenging enough to require sustained effort, but also reachable for that user given their background, role, existing files, and available collaborators.

The objectives are expressed as several deliverable work packages. Each deliverable specifies a concrete professional outcome, the expected progress over the simulated period, and the output artifacts to be created or revised. For example, a deliverable may require an analysis workbook, an internal memo, a presentation deck, supporting materials, and a final PDF package. Some deliverables may also depend on others, reflecting how real productivity work often unfolds through connected projects rather than independent one-off tasks.

\begin{tcolorbox}[
    colback=blue!5,           
    colframe=blue!50!black,   
    arc=2mm,
    boxrule=0.4pt,
    left=3mm, right=3mm, top=2mm, bottom=2mm,
    title=\textbf{Objectives for a Month of Work},
    fonttitle=\small\bfseries,
    fontupper=\footnotesize,
    breakable
]
\setlength{\parskip}{4pt}
\setlength{\parindent}{0pt}

\textbf{Simulation period:} 2026-01-05 to 2026-01-30 (20 working days, 5 deliverables).

\bigskip

{\color{derivecolor}\rule[-0.4em]{2pt}{1.2em}}\hspace{0.5em}\textbf{Deliverable 1: 2026 VCMM Model Portfolio Refresh} \hfill \emph{Target: 2026-01-28}

\smallskip

Finalize the firm's three model portfolios (Conservative, Balanced, Growth) using Vanguard's December 2025 VCMM 10-year projections, author the 2026 Capital Markets Outlook, and present the refreshed strategic asset allocation to the Investment Committee for formal adoption. Justify any weight changes $>$100\,bps versus 2025 allocations.

\smallskip

\hangindent=2em\hangafter=0
\textit{Milestones.} \textbf{W1} obtain VCMM 2026 dataset, draft delta table\,$\to$ \textbf{W2} rebuild allocation models v3\,$\to$ \textbf{W3} complete Capital Markets Outlook\,$\to$ \textbf{W4} present to IC, finalize.\par

\hangindent=2em\hangafter=0
\textit{Expected Artifacts (8 total):}\\
\hspace*{2em}\texttt{\small D:/ModelPortfolios/VCMM\_2026/AllocationModel\_Conservative\_v3.xlsx}\\
\hspace*{2em}\texttt{\small D:/ModelPortfolios/VCMM\_2026/CapMarketsOutlook\_2026\_FINAL\_v3.pdf}\\
\hspace*{2em}\emph{\ldots and 6 more.}

\bigskip

{\color{derivecolor}\rule[-0.4em]{2pt}{1.2em}}\hspace{0.5em}\textbf{Deliverable 2: Robert Castellano HNW Onboarding Package} \hfill \emph{Target: 2026-01-29}

\smallskip

Complete the onboarding work package for new HNW client Robert Castellano (\$7.2M investable, retired tech executive, conservative-to-moderate risk tolerance, 20-year retirement horizon). Requires a follow-up discovery call, completed risk questionnaire, client-specific IPS, and a proposed allocation calibrated to the refreshed 2026 model (depends on Deliverable 1). Compliance sign-off by James Whitfield required before delivery.

\smallskip

\hangindent=2em\hangafter=0
\textit{Milestones.} \textbf{W1} discovery call, risk questionnaire\,$\to$ \textbf{W2} draft IPS v1, initial allocation\,$\to$ \textbf{W3} refresh allocation, run Monte Carlo\,$\to$ \textbf{W4} compliance review, deliver binder.\par

\hangindent=2em\hangafter=0
\textit{Expected Artifacts (8 total):}\\
\hspace*{2em}\texttt{\small D:/ClientWork/Castellano\_Robert/IPS\_Castellano\_FINAL.pdf}\\
\hspace*{2em}\texttt{\small D:/ClientWork/Castellano\_Robert/Kickoff\_Presentation\_Castellano.pptx}\\
\hspace*{2em}\emph{\ldots and 6 more.}

\bigskip

{\color{derivecolor}\rule[-0.4em]{2pt}{1.2em}}\hspace{0.5em}\textbf{Deliverables 3--5} (summarized)

\smallskip

\hangindent=2em\hangafter=1
\textbf{D3.} \emph{Systematic Rebalancing Trigger Framework v3} --- finalize the quantitative rebalancing tool after Sandra Okonkwo's peer review; combine VCMM forward-return differentials with drift thresholds; pilot on three live portfolios. \emph{Target: 2026-01-30.}\par

\hangindent=2em\hangafter=1
\textbf{D4.} \emph{Alternatives Integration --- Final IC Recommendation} --- convert v2 draft into board-quality recommendation on whether Balanced/Growth portfolios should hold a 5--10\% alternatives sleeve. Presented alongside D1. \emph{Target: 2026-01-28.}\par

\hangindent=2em\hangafter=1
\textbf{D5.} \emph{ESG Equity Overlay --- Final Recommendation} --- firm-wide compliance-approved recommendation on Sustainalytics-screened ESG overlays, including suitability memo and IPS addendum disclosure. \emph{Target: 2026-01-30.}\par

\end{tcolorbox}

\subsubsection{Collaboration Setup}
\label{subsubsec:collaboration_setup}

We then introduce a novel collaboration setting around those objectives. This is a key difference between our simulations and standalone productivity tasks: instead of assuming that all necessary information is given upfront and that the work agent can complete the task independently, we model productivity work as a process that may require coordination with other people. In realistic long-horizon productivity work, the user often needs to clarify requirements, request data or source materials, understand stakeholder preferences, and revise artifacts based on feedback.

To support this, the setup agent creates a small set of simulated collaborators who may participate in the work, such as managers, peers, clients, compliance officers, or external partners. Each collaborator is assigned a role, background, communication style, relevant knowledge, and, when appropriate, private reference materials that can be shared with the user through later collaboration when needed:

\begin{tcolorbox}[
    colback=blue!5,           
    colframe=blue!50!black,   
    arc=2mm,
    boxrule=0.4pt,
    left=3mm, right=3mm, top=2mm, bottom=2mm,
    title=\textbf{Simulated Collaborators (Excerpt)},
    fonttitle=\small\bfseries,
    fontupper=\footnotesize,
    breakable
]
\setlength{\parskip}{4pt}
\setlength{\parindent}{0pt}

{\color{derivecolor}\rule[-0.4em]{2pt}{1.2em}}\hspace{0.5em}\textbf{David Hartley} \hfill \emph{Manager (Managing Director)}

\smallskip

\textit{Background.} 58 years old, MBA Wharton, CFA charterholder. Runs Meridian's Denver office (12 advisor teams, \$2.1B firm-wide AUM) and chairs the Denver Investment Committee. Numbers-first, skeptical of narrative arguments, demands complete staff work and clear executive summaries.

\textit{Communication style.} Terse. Email subjects like ``VCMM refresh --- 3 items.'' Response latency 24--48 hours. Prefers bullet points; ignores long prose.

\textit{Reference files (held privately).}\\
\hspace*{1.5em}\texttt{\small DirectorExpectations\_2026\_Priorities.docx} --- Hartley's written 2026 priorities for Margaret's team; KPIs include ``any model change $>$150\,bps requires sensitivity analysis.''\\
\hspace*{1.5em}\texttt{\small Hartley\_RedlineSample\_2025\_CapMarketsOutlook.docx} --- exemplar redline showing his editing standards (active voice, source citations, no hedging verbs).\\
\hspace*{1.5em}\texttt{\small IC\_ReviewChecklist\_ModelPortfolio\_Refresh.docx} --- 8-item checklist for evaluating annual model refreshes.

\bigskip

{\color{derivecolor}\rule[-0.4em]{2pt}{1.2em}}\hspace{0.5em}\textbf{Robert Castellano} \hfill \emph{Client (new HNW onboarding)}

\smallskip

\textit{Background.} 67 years old, retired tech executive. \$7.2M investable across Schwab brokerage, Vanguard IRA, Roth IRA, and a taxable joint account with his wife Elaine. Conservative-to-moderate risk tolerance, 20-year retirement horizon, \$280k annual living budget. Plans an Aspen second-home purchase Q3 2026 requiring \$800k--\$1.0M in liquidity. Holds a concentrated CSCO position (4{,}200 shares, low cost basis).

\textit{Communication style.} Polite, detailed, asks technical follow-up questions. Will read every line of an IPS and mark up anything unclear. Will push back on any recommendation that doesn't explicitly account for the Aspen liquidity event.

\textit{Reference files (held privately).}\\
\hspace*{1.5em}\texttt{\small Castellano\_AccountStatements\_YE2025.xlsx} --- consolidated YE2025 holdings across all four accounts; \emph{contains a 1.7\% allocation discrepancy between summary and position-level rows that the agent must detect}.\\
\hspace*{1.5em}\texttt{\small Castellano\_RiskQuestionnaire\_Completed.pdf} --- score 63/100 (conservative-moderate); notable preference for simpler products.\\
\hspace*{1.5em}\texttt{\small Castellano\_DiscoveryCall\_Followup\_Questions.docx} --- seven written questions including Roth conversion, CSCO concentration, and CPA coordination.\\
\hspace*{1.5em}\texttt{\small Castellano\_SpouseNotes\_Elaine.docx} --- notes on co-signatory Elaine's preferences (more conservative, against illiquid investments).

\bigskip

{\color{derivecolor}\rule[-0.4em]{2pt}{1.2em}}\hspace{0.5em}\textbf{Sandra Okonkwo} \hfill \emph{Peer (Senior Advisor \& IC Member)}

\smallskip

\textit{Background.} 49 years old, CFA, fixed-income and LDI specialist. Manages \$285M AUM. Direct and technically rigorous; will not sign off on anything she has not personally reproduced. Reviews Margaret's rebalancing tool; Q4 2025 peer review flagged three issues v3 must address: tax-lot awareness, cash-drag treatment, crisis-regime correlation breakdown.

\textit{Communication style.} Direct, technical. Responds within hours. Pushes back in writing when math doesn't work. Prefers Excel attachments with working formulas, not values-only.

\textit{Reference files (held privately).}\\
\hspace*{1.5em}\texttt{\small Sandra\_PeerReview\_RebalancingTool\_v2.docx} --- written peer review with specific cell references and a proposed tax-lot scoring formula.\\
\hspace*{1.5em}\texttt{\small Sandra\_HistoricalRebalancingCalls\_2024-2025.xlsx} --- log of her 23 actual rebalancing decisions; v3 expected to flag $\geq$18 in retrospective validation.\\
\hspace*{1.5em}\texttt{\small Sandra\_CrisisRegime\_Correlations\_2020Q1\_2022.xlsx} --- crisis-period correlation matrices for stress-testing v3.

\bigskip

{\color{derivecolor}\rule[-0.4em]{2pt}{1.2em}}\hspace{0.5em}\textbf{Other collaborators} (summarized)

\smallskip

\hangindent=2em\hangafter=1
\textbf{Nathaniel Ortiz} --- \emph{CIO and IC voting member.} Bogle-influenced, wary of complexity. Evaluates the alternatives recommendation (D4) using his ``Why Not the Simplest Thing'' framework. Communicates minimally; writes one-word marginalia (``unpersuasive'', ``asserted'').\par

\hangindent=2em\hangafter=1
\textbf{James Whitfield} --- \emph{Compliance Officer.} Former SEC examiner. Authority over IPS and ESG suitability under SEC Marketing Rule 206(4)-1 and 2024 DOL guidance. Documents everything; cites manual sections in feedback.\par

\hangindent=2em\hangafter=1
\textbf{Patricia Huang} --- \emph{External (Vanguard).} Provides the official VCMM 2026 dataset, methodology white paper, and alternatives supplement. Pure data provider; will not answer client-specific allocation questions.\par

\hangindent=2em\hangafter=1
\textbf{Kevin Tran} --- \emph{Direct report (junior associate).} Performs Morningstar/Bloomberg data pulls and first-pass spreadsheet assembly. Eager but occasionally sloppy --- \emph{intentionally introduces unit-mismatch errors (bps vs.\ \%) and missing data-as-of dates that the agent must catch}.\par

\end{tcolorbox}

\subsection{Planning and Daily Work Simulation}
\label{subsec:planning_daily_work_simulation}

After setup, the work agent acts as the computer's user and carries out the productivity objectives over the simulated period. The agent is given the user profile, the productivity objectives, the simulated collaborator descriptions, and access to the synthetic computer. It does not see collaborators' private reference materials unless they are later shared through collaboration. To complete the deliverable work packages, the agent must operate over the computer: navigating the filesystem for grounding, reading existing artifacts, creating or revising files, communicating with simulated collaborators, and using later replies to continue the work.

The simulation proceeds in weekly and daily cycles. At the beginning of each week, the work agent creates a plan for the coming workdays, based on the productivity objectives, remaining deliverables, prior progress, current computer state, and expected collaboration needs. The plan breaks the week into daily activities, such as deep work, review, administrative cleanup, and outreach. Each activity specifies the files to create or modify, the source artifacts to consult, and any collaborators to contact:
\begin{tcolorbox}[
    colback=blue!5,
    colframe=blue!50!black,
    arc=2mm,
    boxrule=0.4pt,
    left=3mm, right=3mm, top=2mm, bottom=2mm,
    title=\textbf{Weekly Plan --- Week 1 (Excerpt)},
    fonttitle=\small\bfseries,
    fontupper=\footnotesize,
    breakable
]
\setlength{\parskip}{4pt}
\setlength{\parindent}{0pt}

\textbf{Week of 2026-01-05 to 2026-01-09} \quad \emph{Focus: data intake and scoping --- secure VCMM 2026 dataset, Sustainalytics refresh, Castellano discovery, Sandra's peer review.}

\bigskip

{\color{derivecolor}\rule[-0.4em]{2pt}{1.2em}}\hspace{0.5em}\textbf{Monday, 2026-01-05}

\smallskip

\hangindent=4em\hangafter=1
\textbf{10:30 \emph{(outreach)}} \quad Email Patricia Huang requesting (a) VCMM 2026 Q1 asset-class projections (Jan 6 release), (b) VCMM 2026 Alternatives Supplement, (c) 2026 Methodology White Paper. \\ \emph{$\to$ Outreach to:} Patricia Huang \textbar{} \emph{Deliverable:} D1.\par

\hangindent=4em\hangafter=1
\textbf{14:00 \emph{(outreach)}} \quad Email Robert Castellano to schedule follow-up discovery call (Wednesday Jan 7 afternoon); attach agenda; request year-end account statements and completed risk questionnaire in advance. \\ \emph{$\to$ Outreach to:} Robert Castellano \textbar{} \emph{Deliverable:} D2.\par

\hspace*{4em}\emph{$\vdots$ \quad (5 more activities: morning market review, 1:1 with Hartley, outreach to Sandra, Kevin, and Whitfield)}

\bigskip

{\color{derivecolor}\rule[-0.4em]{2pt}{1.2em}}\hspace{0.5em}\textbf{Tuesday, 2026-01-06}

\smallskip

\hangindent=4em\hangafter=1
\textbf{10:30 \emph{(review)}} \quad Read Castellano reference files in preparation for Wednesday's discovery call; prepare structured call agenda flagging the Schwab international allocation discrepancy (18\% stated vs.\ 16.3\% calculated). \\
\emph{Creates:} \filepath{D:/ClientWork/Castellano\_Robert/DiscoveryCall\_Prep\_2026-01-07.docx} \\
\emph{Derived from:} \filepath{Castellano\_AccountStatements\_YE2025.xlsx}, \emph{\ldots and 3 more} \\
\emph{Deliverable:} D2.\par

\hspace*{4em}\emph{$\vdots$ \quad (4 more activities: morning review, Sustainalytics data request, client call block, Hartley reference file study)}

\bigskip

{\color{derivecolor}\rule[-0.4em]{2pt}{1.2em}}\hspace{0.5em}\textbf{Wednesday, 2026-01-07}

\smallskip

\hangindent=4em\hangafter=1
\textbf{11:00 \emph{(admin)}} \quad Wednesday IC huddle with Sandra Okonkwo and David Hartley: confirm Jan 28 IC agenda; review v2 peer review status and v3 scope; brief on Castellano timeline and ESG compliance framework.\par

\hangindent=4em\hangafter=1
\textbf{13:00 \emph{(deep work)}} \quad Conduct follow-up discovery call with Robert Castellano ($\sim$60 min). Write up comprehensive notes covering account structure, liquidity and tax constraints, income requirements, investment preferences, and action items. \\
\emph{Creates:} \filepath{D:/ClientWork/Castellano\_Robert/DiscoveryCall\_Notes\_2026-01-07.docx} \\
\emph{Derived from:} \filepath{Castellano\_AccountStatements\_YE2025.xlsx}, \emph{\ldots and 3 more} \\
\emph{Deliverable:} D2.\par

\hangindent=4em\hangafter=1
\textbf{15:30 \emph{(review)}} \quad Review Sandra's Q4 2025 peer review documents; catalogue the three outstanding issues (tax-lot awareness, cash-drag treatment, crisis-regime correlations) with precise scope; estimate v3 rework for week 2. \\
\emph{Creates:} \filepath{D:/ModelPortfolios/RebalancingFramework/PeerReview\_IssuesCatalogue\_v2\_2026-01-07.docx} \\
\emph{Derived from:} \filepath{Sandra\_PeerReview\_RebalancingTool\_v2.docx}, \emph{\ldots and 2 more} \\
\emph{Deliverable:} D3.\par

\hspace*{4em}\emph{$\vdots$ \quad (1 more activity: morning market review)}

\bigskip

{\color{derivecolor}\rule[-0.4em]{2pt}{1.2em}}\hspace{0.5em}\textbf{Thursday, 2026-01-08}

\smallskip

\hangindent=4em\hangafter=1
\textbf{10:30 \emph{(deep work)}} \quad Build the VCMM 2026 vs.\ 2025 delta analysis spreadsheet (5-tab Excel workbook): summary delta table across 12 asset classes; per-asset-class detail; flagged material changes ($|\Delta|>50$\,bps); correlation matrix changes; workings and sources. \\
\emph{Creates:} \filepath{D:/ModelPortfolios/VCMM\_2026/VCMM\_2026\_vs\_2025\_DeltaAnalysis.xlsx} \\
\emph{Derived from:} \filepath{VCMM\_2026\_AssetClassProjections\_Release.xlsx}, \emph{\ldots and 2 more} \\
\emph{Deliverable:} D1.\par

\hangindent=4em\hangafter=1
\textbf{16:00 \emph{(email)}} \quad Send Kevin written feedback on the Morningstar alt funds export: three expense ratios entered as bps instead of \%; missing data-as-of date in header. Request corrected version by Friday EOD. \\ \emph{$\to$ Outreach to:} Kevin Tran \textbar{} \emph{Deliverable:} D4.\par

\hspace*{4em}\emph{$\vdots$ \quad (4 more activities: morning review, VCMM intake reading, client call block, etc.)}

\bigskip

{\color{derivecolor}\rule[-0.4em]{2pt}{1.2em}}\hspace{0.5em}\textbf{Friday, 2026-01-09}

\smallskip

\hangindent=4em\hangafter=1
\textbf{11:00 \emph{(deep work)}} \quad Write Preliminary Findings Memo to David Hartley summarizing week 1 VCMM intake and key model implications (5--7 pages, with executive summary, delta table, material changes for IC, preliminary allocation implications, and open items requiring Hartley's direction). \\
\emph{Creates:} \filepath{D:/ModelPortfolios/VCMM\_2026/PreliminaryFindings\_Memo\_Week1\_2026-01-09.docx} \\
\emph{Derived from:} \filepath{VCMM\_2026\_vs\_2025\_DeltaAnalysis.xlsx}, \emph{\ldots and 2 more} \\
\emph{Deliverable:} D1.\par

\hangindent=4em\hangafter=1
\textbf{14:00 \emph{(outreach)}} \quad Email David Hartley attaching the Preliminary Findings Memo. Subject: ``VCMM 2026 --- Prelim findings + 3 items for your direction.'' Three-bullet summary in body; request response by Tuesday Jan 13. \\ \emph{$\to$ Outreach to:} David Hartley \textbar{} \emph{Deliverable:} D1.\par

\hspace*{4em}\emph{$\vdots$ \quad (4 more activities: morning review, Kevin drift data review, weekly coaching note, week wrap-up)}

\end{tcolorbox}

After the weekly plan is created, the work agent executes it one workday at a time. Each day is run as a separate agent session. At the start of each day, the agent restores the current work context by reviewing the activity log, checking the current computer state, and reading any new collaborator replies or shared files from previous days. It then performs the day's planned activities by reading relevant existing artifacts, creating or revising the required output files, and sending messages or shared files to collaborators when needed. At the end of the day, the simulation records the new files, revised artifacts, collaborator interactions, and activity history, so that the next daily session can continue from the updated computer state.

\begin{tcolorbox}[
    colback=blue!5,
    colframe=blue!50!black,
    arc=2mm,
    boxrule=0.4pt,
    left=3mm, right=3mm, top=2mm, bottom=2mm,
    title=\textbf{Daily Activity Log --- 2026-01-07 (Excerpt)},
    fonttitle=\small\bfseries,
    fontupper=\footnotesize,
    breakable
]
\setlength{\parskip}{4pt}
\setlength{\parindent}{0pt}

\textbf{2026-01-07 (Wednesday)} \quad \emph{Activity log entries written by the agent during execution of the day's plan.}

\bigskip

{\color{derivecolor}\rule[-0.4em]{2pt}{1.2em}}\hspace{0.5em}\textbf{11:00} \quad \emph{Investment committee huddle with Sandra Okonkwo and David Hartley.}

\smallskip

\hangindent=2em\hangafter=0
45-minute meeting. Topics covered: (i) IC Jan 28 agenda confirmed --- VCMM refresh and alternatives recommendation as primary items, ESG overlay compliance status as third item pending Whitfield sign-off; (ii) rebalancing v3 scope confirmed: 4 issues in scope (tax-lot scoring, cash-drag as separate sleeve, crisis-regime correlations, drift-band asymmetry); prototype due Jan 20; (iii) Castellano timeline: discovery call this afternoon, IPS v1 target Jan 30, IPS review meeting Jan 29, Whitfield compliance review Feb 6; (iv) ESG framework: Whitfield involvement confirmed. Action items: Margaret to confirm v3 development timeline by Friday Jan 9; Hartley to send IC pre-read list by Jan 20.\par

\bigskip

{\color{derivecolor}\rule[-0.4em]{2pt}{1.2em}}\hspace{0.5em}\textbf{14:00} \quad \emph{Castellano follow-up discovery call.}

\smallskip

\hangindent=2em\hangafter=0
$\sim$60 minutes (2:00--3:05pm MT). Attendees: Robert Castellano, Elaine Castellano (joined for final 15 min), Margaret Forsythe. Key outcomes: confirmed all four account structures (\$4.3M Schwab, \$1.9M Vanguard IRA, \$420K Roth, \$580K taxable joint); \textbf{confirmed Schwab international discrepancy --- 16.3\% position-level vs.\ 18\% summary, Castellano independently noticed this}; confirmed Aspen liquidity source restricted to Schwab/taxable only (Roth excluded); confirmed CSCO phased disposition 2026--2027; confirmed Roth conversion interest $\sim$\$200K/year in 2026--2031 window with CSCO gain-stacking sequencing concern; \$15K/month net withdrawal requirement from Elaine; CPA contact with Marcus Lin authorized.\par

\bigskip

{\color{derivecolor}\rule[-0.4em]{2pt}{1.2em}}\hspace{0.5em}\textbf{15:30} \quad \emph{Wrote formal discovery call notes.}

\smallskip

\hangindent=2em\hangafter=0
Created \filepath{D:/ClientWork/Castellano\_Robert/DiscoveryCall\_Notes\_2026-01-07.docx} (5--7 pages, 6 sections): (1) Account Structure Confirmed --- 16.3\% corrected international figure documented, 1.7\% cash misclassification flagged; (2) Liquidity and Tax Constraints --- Aspen \$800K--\$1.0M from Schwab/taxable only, CSCO \$84K gain phased 2026--2027, Roth conversion \$200K/yr with CSCO sequencing risk flagged; (3) Income and Withdrawal --- \$280K gross / \$200K after-tax / Elaine fixed \$15K/month; (4) Investment Preferences --- 6/10 risk tolerance, plain index fund preference, Elaine firm against illiquid/alternatives; (5) 7 pending items; (6) 8 action items with owners and dates. Includes Whitfield-required elements: CSCO concentration acknowledgment, Aspen liquidity documentation, Roth conversion disclosure notes.\par

\bigskip

{\color{derivecolor}\rule[-0.4em]{2pt}{1.2em}}\hspace{0.5em}\textbf{17:00} \quad \emph{Reviewed Sandra's peer review files; wrote issues catalogue.}

\smallskip

\hangindent=2em\hangafter=0
Read three peer review files received Jan 6 (\filepath{Sandra\_PeerReview\_RebalancingTool\_v2.docx}, \filepath{Sandra\_HistoricalRebalancingCalls\_2024-2025.xlsx}, \filepath{Sandra\_CrisisRegime\_Correlations\_2020Q1\_2022.xlsx}). Created \filepath{D:/ModelPortfolios/RebalancingFramework/PeerReview\_IssuesCatalogue\_v2\_2026-01-07.docx} (3--4 pages) cataloguing four issues. \emph{Issue 1 (tax-lot awareness):} scoring formula \emph{tax-cost penalty} $=$ unrealized gain/market value $\times$ marginal rate; lots $>$20\% gain/MV get 50\% sell-priority discount. \emph{Issue 2 (cash-drag):} separate sleeve row, 2--5\% target, $\pm$200\,bps threshold. \emph{Issue 3 (crisis correlations):} equity/bond correlation flips from $-0.20$ to $+0.45$ (2020-Q1) and $+0.62$ (2022); REIT/equity spikes to 0.78. \emph{Issue 4 (drift-band asymmetry):} asymmetric bands per risk tier. Week 2 development plan with Jan 20 prototype delivery to Sandra.\par

\bigskip

\hspace*{2em}\emph{$\vdots$ \quad (2 more entries: 08:00 morning market review with \filepath{ResearchLog\_2026.docx} update; 08:30 filing of Castellano reference documents into \filepath{D:/ClientWork/Castellano\_Robert/})}

\end{tcolorbox}

This weekly planning and daily execution loop repeats until the simulated period is complete. As the simulation proceeds, the synthetic computer is changed by the agent's work: new files are added, existing artifacts are revised, collaborator exchanges are recorded, and the file graph is updated to reflect how outputs build on prior materials. The resulting trajectory provides process-level signals from the agent's planning, grounding, collaboration, and revision behavior, while the final deliverables provide outcome-level signals about whether the productivity objectives were completed successfully.

\section{Experiments}

\subsection{Experimental Setup}
\label{subsec:experimental_setup}

We use the Claude Code SDK as the agent runtime. Unless otherwise noted, the agents are powered by Claude Sonnet 4.6; Claude Opus 4.6 is used for long-horizon simulation setup (Section~\ref{subsec:setup}). For artifact creation, the agent is equipped with external skills: we use Anthropic's skills for non-Office artifact types, and we use MiniMax's open-source skills for Office-related artifacts---\texttt{minimax-docx}, \texttt{minimax-xlsx}, \texttt{pptx-generator}, and \texttt{minimax-pdf}.

\begin{figure}[htbp]
    \centering
    \begin{minipage}[t]{0.48\linewidth}
        \centering
        \includegraphics[width=\linewidth]{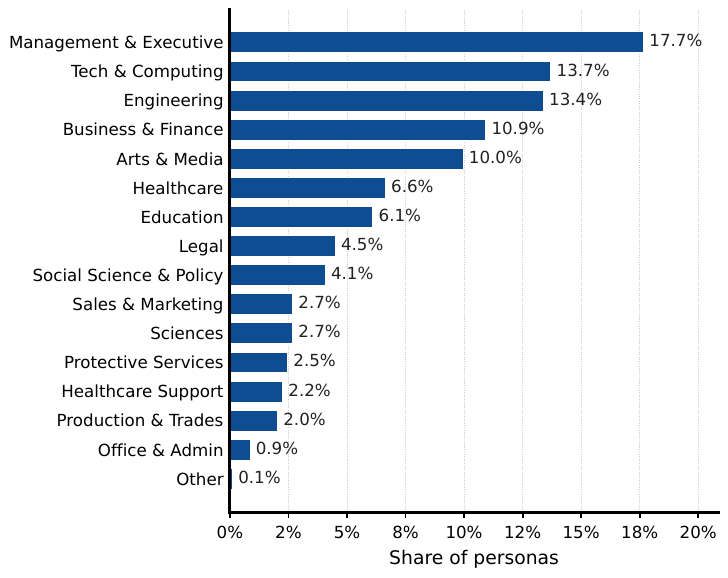}
        \captionof{figure}{Occupation distribution of the 1,000 sampled personas}
        \label{fig:persona_distribution}
    \end{minipage}\hfill
    \begin{minipage}[t]{0.48\linewidth}
        \centering
        \includegraphics[width=\linewidth]{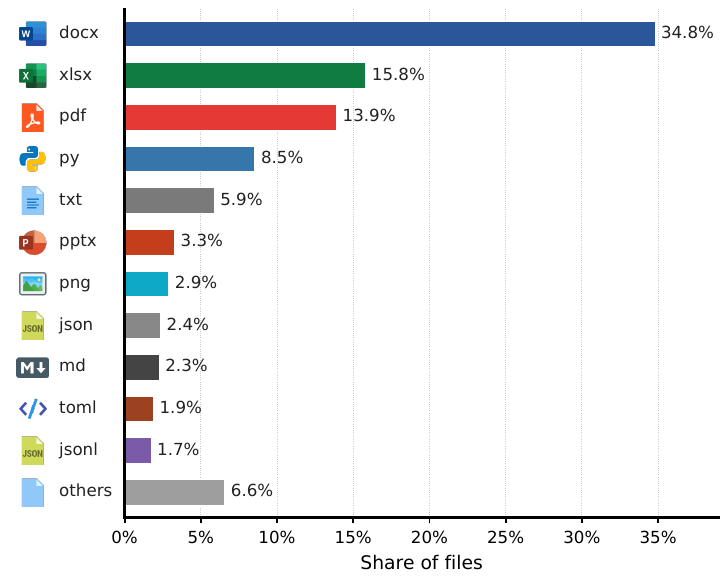}
        \captionof{figure}{Artifact type distribution across the synthetic computers}
        \label{fig:artifact_distribution}
    \end{minipage}
\end{figure}

Following our prior methodology~\citep{ge2024scaling}, we curate a large collection of personas and sample 1,000 personas as the starting points for synthetic computer creation. Their occupation distribution is shown in Figure~\ref{fig:persona_distribution}.

\subsection{Results}
\label{subsec:results}

Table~\ref{tab:synth-pc} summarizes the structural properties of the synthetic computers before and after simulation. Before simulation, each synthetic computer contains about 112 files on average; after a month of simulated work, this increases to about 197 files. Directory counts also increase moderately, while directory depth remains largely stable. This suggests that the work agent extends the existing computer environment mainly by creating and revising artifacts within the established organization, rather than producing unrealistic directory structures.

Figure~\ref{fig:artifact_distribution} shows the artifact type distribution across the synthetic computers. The generated files are dominated by productivity artifacts: DOCX, XLSX, PDF, and PPTX together account for 67.8\% of all files, with DOCX alone contributing 34.8\%. The remaining files include code, text, images, and structured data formats, reflecting supporting materials that commonly appear alongside productivity documents in realistic computer environments.

\begin{figure}[h]   
\centering

\begin{minipage}[h]{0.5\linewidth}
\centering
\captionof{table}{Synthetic computer statistics before and after a month of simulated work}
\label{tab:synth-pc}
\small
\begin{tabular}{@{}lrrrr@{}}
\toprule
                                     & Mean  & Median & Min  & Max  \\
\midrule
\multicolumn{5}{@{}l}{\textit{Files per computer}}                  \\
\quad Pre-simulation                 & 111.6 & 89     & 72   & 595  \\
\quad Post-simulation                & 197.4 & 174    & 131  & 670  \\
\addlinespace
\multicolumn{5}{@{}l}{\textit{Directories per computer}}            \\
\quad Pre-simulation                 & 30.4  & 25     & 16   & 181  \\
\quad Post-simulation                & 36.0  & 29     & 18   & 190  \\
\addlinespace
\multicolumn{5}{@{}l}{\textit{Avg directory depth}}                 \\
\quad Pre-simulation                 & 3.39  & 3.22   & 2.72 & 6.68 \\
\quad Post-simulation                & 3.40  & 3.25   & 2.60 & 6.54 \\
\addlinespace
\multicolumn{5}{@{}l}{\textit{Max directory depth}}                 \\
\quad Pre-simulation                 & 5.5   & 5      & 4    & 13   \\
\quad Post-simulation                & 5.6   & 5      & 4    & 13   \\
\bottomrule
\end{tabular}
\end{minipage}
\hfill
\begin{minipage}[h]{0.45\linewidth}
\centering
\captionof{table}{File sizes by common artifact formats}
\label{tab:office-sizes}
\small
\begin{tabular}{@{}lrrr@{}}
\toprule
       & Mean (KB) & Median (KB) & p95 (KB) \\
\midrule
\multicolumn{4}{@{}l}{\textit{Reference files (held by collaborators)}}  \\
\quad docx   & 26.6   & 37.8   & 43.0   \\
\quad xlsx   & 51.0   & 13.3   & 214.2  \\
\quad pptx   & 277.1  & 250.7  & 843.6  \\
\quad pdf    & 55.8   & 34.1   & 151.8  \\
\addlinespace
\multicolumn{4}{@{}l}{\textit{Final deliverables}}      \\
\quad docx   & 14.8   & 12.0   & 37.5   \\
\quad xlsx   & 28.1   & 16.4   & 74.1   \\
\quad pptx   & 615.4  & 576.6  & 1229.1 \\
\quad pdf    & 141.8  & 88.9   & 359.3  \\
\addlinespace
\multicolumn{4}{@{}l}{\textit{All}}             \\
\quad docx   & 14.6   & 10.8   & 38.6   \\
\quad xlsx   & 75.7   & 15.5   & 104.3  \\
\quad pptx   & 593.7  & 546.5  & 1200.6 \\
\quad pdf    & 383.1  & 83.4   & 1116.8 \\
\bottomrule
\end{tabular}
\end{minipage}
\end{figure}

Table~\ref{tab:office-sizes} reports file sizes for common artifact formats. The generated artifacts are non-trivial in size, especially for presentations and PDFs, both among collaborator-held reference files and final deliverables. These statistics indicate that the simulations produce content-rich productivity artifacts rather than lightweight placeholder files.

We show in Table~\ref{tab:run-telemetry} that each simulation is a substantial long-horizon run. On average, the work agent takes 2,272 turns and 8.59 hours to complete one simulation, with most of the cost coming from daily execution rather than weekly planning. The simulations also involve non-trivial collaboration: each synthetic computer has 5.5 simulated collaborators on average, and the work agent exchanges about 31 communications per simulation. These statistics indicate that the simulations go beyond one-shot task completion, requiring sustained planning, daily execution, and repeated coordination with collaborators.

\begin{table}[t]
\centering
\caption{Work-agent simulation statistics across $1,000$ synthetic computers}
\label{tab:run-telemetry}
\begin{tabular}{@{}lrrrr@{}}
\toprule
                                       & Mean   & Median & Min   & Max   \\
\midrule
\multicolumn{5}{@{}l}{\textit{\# Turns}}              \\
\quad Weekly planning                  & 63     & 63     & 48    & 85    \\
\quad Daily execution                  & 2209   & 2166   & 1499  & 3180  \\
\quad Total                            & 2272   & 2234   & 1551  & 3248  \\
\addlinespace
\multicolumn{5}{@{}l}{\textit{Wall-clock time (hours)}}  \\
\quad Weekly planning                  & 0.59   & 0.55   & 0.38  & 1.00  \\
\quad Daily execution                  & 8.00   & 7.74   & 5.82  & 10.79 \\
\quad Total                            & 8.59   & 8.31   & 6.27  & 11.67 \\
\addlinespace
\multicolumn{5}{@{}l}{\textit{Collaboration}}            \\
\quad \# Simulated collaborators          & 5.5    & 5.0    & 5.0   & 8.0   \\
\quad \# Communications                   & 31     & 30     & 17    & 56    \\
\bottomrule
\end{tabular}
\end{table}

\subsubsection{Final Deliverable Evaluation}\label{subsubsec:final}

We evaluate final deliverables with a rubric created for each evaluated computer and its productivity objectives\footnote{We here introduce a simple rubric-based judge for this technical report. We skip discussing more advanced rubric generation and automatic evaluation methods~\citep{liu2026openrubricsscalablesyntheticrubric,shen2026rethinkingrubricgenerationimproving}, which are orthogonal to the synthetic computer and simulation methodology studied here.}. To make the rubric less tied to a single run, we run the same simulation setting five times. For each run, we ask a judge to inspect the final deliverables and write a draft rubric based on what a good solution should satisfy. The judge is equipped with tools and skills to read the deliverables directly and, when needed, inspect screenshots to assess visual quality. When writing the rubric, the judge also considers the productivity objectives, the expectations for these deliverables, and the preferences or requirements expressed by simulated collaborators, all of which are created by the setup agent as described in Section~\ref{subsec:setup}.

This gives us five draft rubrics for each evaluated computer. We then merge the draft rubrics into a final rubric. Since a rubric from one run may be too narrow, the different runs serve as complementary references for writing a more complete and more general rubric for the same productivity objectives. A typical example is shown below:

\begin{tcolorbox}[
    colback=blue!5,
    colframe=blue!50!black,
    arc=2mm,
    boxrule=0.4pt,
  left=8pt, right=8pt, top=6pt, bottom=6pt,
  title={\small\bfseries Sample Rubric: Sandra Okonkwo on dlv\_003},
  fonttitle=\small,
  enhanced, breakable
]
\small
\noindent
\textbf{Deliverable:} v3 portfolio-rebalancing tool (financial advisor setting) \\ [2pt]
\textbf{Total items:} 55\quad
\textbf{Total points:} 176\quad
\textbf{Source mix:} spec (6), interaction (24), expertise (11), reference (4), quality (10)

\medskip
\noindent\textit{Representative \emph{problem-level} items (4 sources shown):}

\medskip
\noindent\textsf{\textbf{[spec, 4pt]}}\quad
All expected deliverable artifacts exist (model workbook, pilot validation
workbook, methodology memo in editable and PDF form, advisor quick reference,
stress-test workbook) in their appropriate file formats.

\smallskip
\noindent\textsf{\textbf{[spec, 2pt]}}\quad
Advisor quick-reference is a single-page (front/back acceptable) document
suitable for daily monitoring use by a junior analyst.

\smallskip
\noindent\textsf{\textbf{[spec, 3pt]}}\quad
Tool emits an Action / Watch / Hold signal for each pilot portfolio with
clear basis (which rule triggered).

\smallskip
\noindent\textsf{\textbf{[interaction, 5pt]}}\quad
Cash is modeled as its own dedicated sleeve with its own target, drift
threshold, and drag attribution---not folded into a bond-equivalent
calculation.

\smallskip
\noindent\textsf{\textbf{[interaction, 4pt]}}\quad
Cash sleeve has an independent Watch trigger when cash exceeds target by a
defined band, and this independent signal is visible in both portfolio
output and backtest output.

\smallskip
\noindent\textsf{\textbf{[interaction, 5pt]}}\quad
Crisis-regime correlation handling is implemented as a parallel,
simultaneously-displayed sensitivity output (long-run alongside crisis
regimes)---not a user-toggled dropdown that defaults to normal.

\smallskip
\noindent\textsf{\textbf{[interaction, 3pt]}}\quad
Regime-alert trigger is informational only (does NOT modify the
Action / Watch / Hold threshold math); clearly labeled as such in the
model and methodology documentation.

\smallskip
\noindent\textsf{\textbf{[expertise, 3pt]}}\quad
A worked example is shown end-to-end (sample inputs, intermediate
computations, final score, and flag determination) and is reproducible
by hand.

\smallskip
\noindent\textsf{\textbf{[expertise, 4pt]}}\quad
Crisis-regime correlation values are internally consistent across all
deliverables (model crisis-regime tab, stress-test workbook, methodology
memo, and quick reference do not disagree on the same underlying number).

\smallskip
\noindent\textsf{\textbf{[expertise, 5pt]}}\quad
Drift threshold definitions are internally consistent across the model,
the methodology memo (both editable and PDF forms), and the advisor
quick reference---no version drift between artifacts.

\smallskip
\noindent\textsf{\textbf{[quality, 2pt]}}\quad
Methodology memo is readable end-to-end in roughly 20 minutes by a
non-quant advisor (concise, well-structured, no excessive jargon).

\smallskip
\noindent\textsf{\textbf{[quality, 3pt]}}\quad
Data-source registry / provenance section documents source files, tab
names, as-of dates, and version of any external feeds used as inputs.

\smallskip
\noindent\textsf{\textbf{[quality, 2pt]}}\quad
No hard-coded magic numbers in the live model---thresholds and inputs
are clearly separated from formulas and traceable to input cells.

\medskip
\noindent\dotfill\ \emph{42 additional items omitted}\ \dotfill\\[2pt]
\noindent\footnotesize\textit{Note: the four \texttt{[reference]} items
in this rubric are tightly grounded in specific numeric values from the
evaluator's source materials and are omitted from this sample.}
\end{tcolorbox}

\begin{figure}[h]
  \centering
  \includegraphics[width=0.48\linewidth]{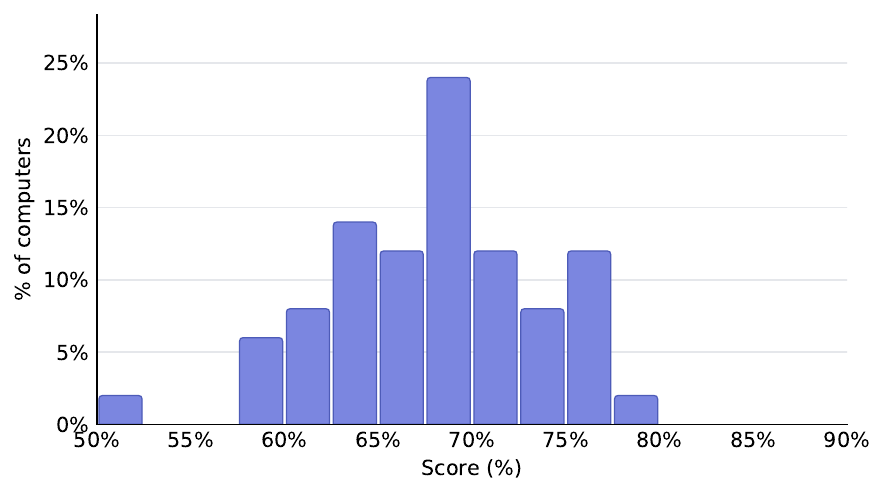}\hfill
  \includegraphics[width=0.48\linewidth]{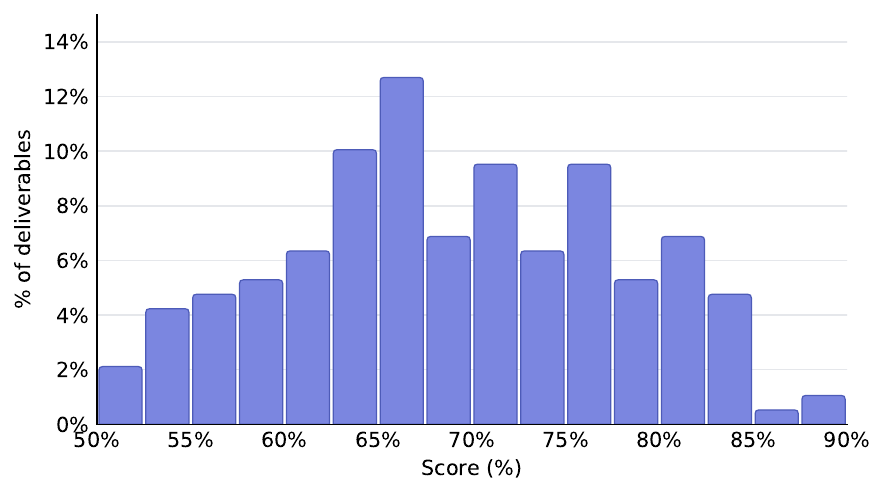}
  \caption{Sore distribution against the rubric.
  \textbf{Left}: per-computer aggregate. \textbf{Right}: per-deliverable.}
  \label{fig:rubric_score_distribution}
\end{figure}

We use a Claude Code SDK agent powered by Claude Opus 4.6 as the judge. Given the final rubric and the final deliverables, the judge scores the deliverables according to the rubric and produces an evaluation summary. We apply this rubric-based evaluation to 100 synthetic computers sampled from the 1,000-computer pool. Figure~\ref{fig:rubric_score_distribution} shows the resulting score distribution. Most scores fall between 60\% and 80\%, suggesting that the work agent can complete many of the required deliverables, while still leaving substantial room for improvement.

\subsubsection{Full Trajectory Analysis}\label{subsubsec:trajectory}

Beyond judging the final deliverables, we analyze the full trajectory of each simulation on its corresponding synthetic computer. The analysis considers both the process and the outcome: how the work agent planned over time, navigated the filesystem, used existing artifacts, coordinated with simulated collaborators, revised intermediate files, and whether the final deliverables satisfied the objectives. For each simulation, we generate a retrospective report that summarizes what the agent did well, where it failed or underperformed, and which mistakes or behaviors are useful as learning signals. We show a representative retrospective report in Appendix \ref{appsec:report}.

This trajectory-level analysis is important because many failures in long-horizon productivity work are not visible from the final files alone. A final deliverable may look acceptable while relying on weak grounding, missing collaborator feedback, or unnecessary rework; conversely, a poor final output may result from an earlier planning or coordination failure. By analyzing the full trajectory, we can extract richer experiential signals about both successful behaviors and failure modes. We evaluate the effectiveness of these signals in Sections~\ref{subsec:in_domain_evaluation} and~\ref{subsec:out_of_domain_evaluation}, where we report in-domain and out-of-domain results, respectively.

\subsection{In-Domain Evaluation}
\label{subsec:in_domain_evaluation}

We evaluate whether the experience extracted from prior simulations can improve the work agent on new synthetic computers from the same distribution (i.e., in-domain evaluation). We split the 1,000 synthetic computers into 900 training\footnote{Here ``training'' refers to the split used for extracting experience and creating skills; we do not update model weights in this evaluation.} computers and 100 held-out test computers. The 900 training computers are used only to extract experience and create skills; the 100 held-out computers are used for evaluation.

From each training simulation, we take the retrospective report described in Section~\ref{subsubsec:trajectory} and extract a set of experience items, such as useful work patterns, lessons, warnings, and common failure modes. We then group these items by the occupation of the computer's user. Within each occupation group, an LLM merges similar items and counts how often each item appears. This gives us a frequency-ranked list of the most common lessons and failure modes for each occupation.

We then give these ranked items to a skill creator\footnote{We use the skill-creator skill by Anthropic: \url{https://github.com/anthropics/skills/blob/main/skills/skill-creator/SKILL.md}}, which writes one occupation-specific skill for each occupation group. The frequency counts help the skill creator focus on the most common and important issues. An example of a resulting skill is shown below:

\begin{tcolorbox}[
    colback=blue!5,
    colframe=blue!50!black,
    arc=2mm,
    boxrule=0.4pt,
    left=8pt, right=8pt, top=6pt, bottom=6pt,
    title={\small\bfseries Sample Skill: \texttt{financial-and-investment-analysts}},
    fonttitle=\small,
    enhanced, breakable
]
\small
\noindent
\textbf{Trigger scope:} Financial models, DCF/LBO/M\&A models, fund reporting,
derivatives pricing, GIPS compliance, SEC filings, pitch books, due diligence,
and any multi-document financial deliverable package.

\medskip
\noindent\textit{Skill is organized into four major sections; representative rules shown.}

\medskip
\noindent\textbf{\S1\; Data Integrity and Single-Source-of-Truth}
\smallskip

\noindent\textsf{\textbf{[source-of-truth]}}\quad
Every shared figure has exactly one authoritative source---the Excel model or
registry. Memos, decks, and PDFs are read-only consumers. Re-open the source
model immediately before composing any section that references it; the gap
between ``I know the WACC is 9.2\%'' and pasting the actual linked value is
where stale figures survive.

\smallskip
\noindent\textsf{\textbf{[registry]}}\quad
For every shared figure record: parameter name, source, exact value, time scope
(quarterly / annual / TTM / LTM), currency, and date confirmed.
A quarterly and an annual figure for the same metric are \emph{not}
interchangeable even when similar in magnitude.

\smallskip
\noindent\textsf{\textbf{[derivation chain]}}\quad
Explicitly confirm whether a collaborator-supplied projection is revenue or
EBITDA and show the full derivation: stated figure $\to$ interpretation $\to$
derived EBITDA at margin $\to$ implied valuation. Confusing the two produces
valuations off by $5$--$10\times$.

\smallskip
\noindent\textsf{\textbf{[methodology change]}}\quad
Any methodology change between draft and final must be propagated to every
dependent file with a change log noting what changed, why, and which files were
updated. A silent change that makes the model, memo, and deck contradict each
other is among the most damaging errors in institutional finance.

\medskip
\noindent\textbf{\S2\; Model Construction and Validation}
\smallskip

\noindent\textsf{\textbf{[sensitivity integrity]}}\quad
Before delivering any sensitivity table, verify it works: adverse scenarios
produce worse outcomes than base; larger moves produce proportionally larger
impacts. Inverted ordering or flat results indicate a model error, not a
formatting choice.

\smallskip
\noindent\textsf{\textbf{[fund return labeling]}}\quad
Gross IRR is pre-fees, pre-carry, pre-expenses; net IRR is after management
fees, carry, and fund expenses. Never present one without labeling which layer;
side-by-side gross/net is standard for IC materials. Subscription-line-inflated
IRR without disclosure is a recurring source of LP dispute.

\smallskip
\noindent\textsf{\textbf{[closed-system check]}}\quad
Verify all valuation inputs (risk-free rate, beta, ERP, cost of equity,
terminal growth) form a mathematically closed system reproducing stated outputs
when computed mechanically. A reviewer must reproduce every WACC component
without undisclosed bridging adjustments.

\medskip
\noindent\textbf{\S3\; Document Hierarchy and Workflow Gates}
\smallskip

\noindent\textsf{\textbf{[version discipline]}}\quad
Once a version is shared it becomes the authentic record of that review cycle.
Never reuse a version number: if released v1.1 has an error, the fix is v1.2.
Put the version string in the \emph{filename}, not just the header.

\smallskip
\noindent\textsf{\textbf{[pre-submission sweep]}}\quad
Use explicit searchable markers---\texttt{[VERIFY]}, \texttt{[TBD]},
\texttt{[PENDING AUDIT]}---during drafting rather than plausible estimates.
A plausible wrong number is more dangerous than a visible gap. Run a mechanical
\texttt{grep} across every deliverable before transmission; unresolved markers
are delivery blockers.

\medskip
\noindent\textbf{\S4\; Regulatory, Compliance, and Certification Standards}
\smallskip

\noindent\textsf{\textbf{[GIPS / net performance]}}\quad
Any document presenting gross performance must also present net-of-fee with
equal prominence---a regulatory requirement under SEC Rule 206(4)-1, not a
stylistic choice. Applies across pitch books, quarterly reports, and LP letters.

\smallskip
\noindent\textsf{\textbf{[cross-engagement isolation]}}\quad
A client name, CUSIP, or fee term from Fund~A appearing in Fund~B's deliverable
is a reportable confidentiality breach. Before transmitting any deliverable,
extract every capitalized multi-word token and confirm each appears on the
engagement's allow-list.

\medskip
\noindent\dotfill\ \emph{Remaining rules omitted}\ \dotfill
\end{tcolorbox}

Finally, we evaluate on the 100 held-out synthetic computers. For each held-out computer, we compare the baseline work agent against the same work agent equipped with the generated occupation-specific skill. Both agents use the same setup and operate on the same synthetic computer; the only difference is whether the work agent has access to the skill extracted from the 900 training simulations. This evaluates whether simulation-derived experience can improve performance on new long-horizon productivity simulations from the same distribution.

\begin{figure}[h]
  \centering
  \begin{minipage}[t]{0.44\textwidth}
    \vspace{0pt}
    \centering
    \small
    \captionof{table}{Effect of occupation skills (i.e., experiential learning signals) on the simulation, measured across $100$ test computers. Each computer is scored using the rubric-based approach discussed in Section~\ref{subsubsec:final}.}
    \label{tab:occ-skill-effect}
    \vspace{2pt}
    \begin{tabular}{@{}lr@{}}
    \toprule
                                                  & Value          \\
    \midrule
    \multicolumn{2}{@{}l}{\textit{Mean score}}                          \\
    \quad Baseline                                    & 61.6\%         \\
    \quad Skill-augmented                             & 68.6\%         \\
    \quad $\Delta$ (Skill-augmented $-$ Baseline)     & +7.0\,pp    \\
    \addlinespace
    \multicolumn{2}{@{}l}{\textit{Per-computer outcome (out of 100)}}   \\
    \quad Skill-augmented better                      & 83 (83\%)      \\
    \quad Baseline better                             & 17 (17\%)      \\
    \bottomrule
    \end{tabular}
  \end{minipage}\hfill
  \begin{minipage}[t]{0.54\textwidth}
    \vspace{0pt}
    \centering
    \includegraphics[width=\linewidth]{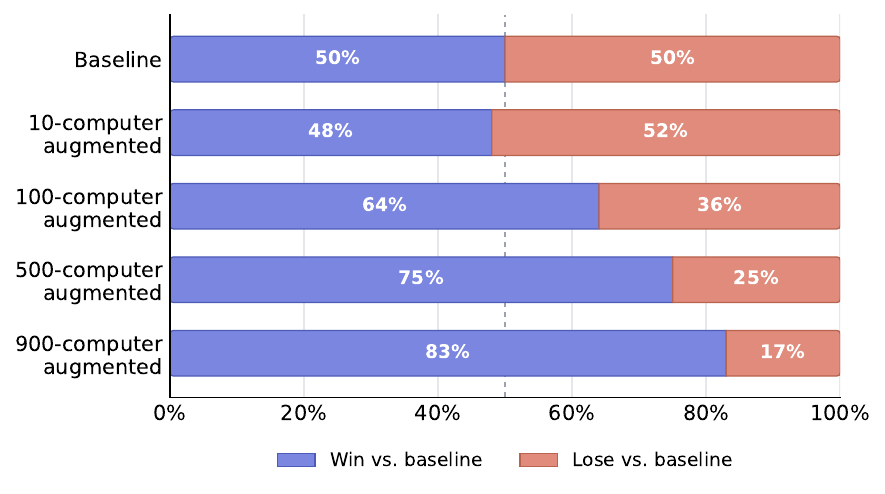}   \captionsetup{skip=-2pt}
    \captionof{figure}{Per-computer win/lose share of skill-augmented variants against the baseline as the number of training computers $N$ grows. Each row aggregates 100 paired comparisons.}
    \label{fig:win-lose-vs-baseline}
  \end{minipage}
\end{figure}

Table~\ref{tab:occ-skill-effect} shows that the occupation skills extracted from the 900 training simulations substantially improve the work agent on the 100 held-out test computers. The mean rubric score increases from 61.6\% to 68.6\%. In paired comparison, the skill-augmented agent outperforms the baseline on 83 out of 100 test computers. This suggests that the trajectory-derived skills capture useful experience that transfers to new synthetic computers from the same distribution.

We show how the effect changes with the number of training computers used to create the skills in Figure~\ref{fig:win-lose-vs-baseline}. With only 10 training computers, the skill-augmented agent does not improve over the baseline, likely because the training set covers too few occupations. For many test computers, no matching occupation skill is available, so the agent may use a weakly related skill that does not fit the user's work context and can even hurt performance. As the number of training computers increases, performance improves for two reasons. First, occupation coverage becomes broader, so test computers are more likely to have a relevant skill. Second, the frequency estimates become more reliable, allowing the skill creator to better identify the common lessons and failure modes that should receive higher priority. Skills extracted from 100, 500, and 900 computers win on 64\%, 75\%, and 83\% of the test computers, respectively. This scaling trend suggests that broader simulation coverage leads to more useful and better-targeted experiential skills.

\subsection{Out-of-Domain Evaluation}
\label{subsec:out_of_domain_evaluation}

\begin{figure}[h]
  \centering
  \begin{minipage}[t]{0.5\textwidth}
    \vspace{0pt}
    \centering
    \captionof{table}{Comparison between our simulations and the GDPVal gold set. Average turns and wall-clock time are measured by running Claude Sonnet 4.6 with the Claude Code SDK.}
    \label{tab:gdpval_task_comparison}
    \vspace{2pt}
    \small
    \begin{tabular}{lcc}
    \toprule
     & Ours & GDPVal \\
    \midrule
    Avg. \# reference files  & 13.8        & 1.18        \\
    Avg. \# computer files   & 112        & 0         \\
    Avg. \# deliverables     & 4.09       & 1.63        \\
    Avg. \# turns            & 2272       & 31        \\
    Avg. wall-clock time     & 8.59 hours & 17 minutes \\
    \bottomrule
    \end{tabular}
  \end{minipage}\hfill
  \begin{minipage}[t]{0.47\textwidth}
    \vspace{0pt}
    \centering
    \includegraphics[width=\linewidth]{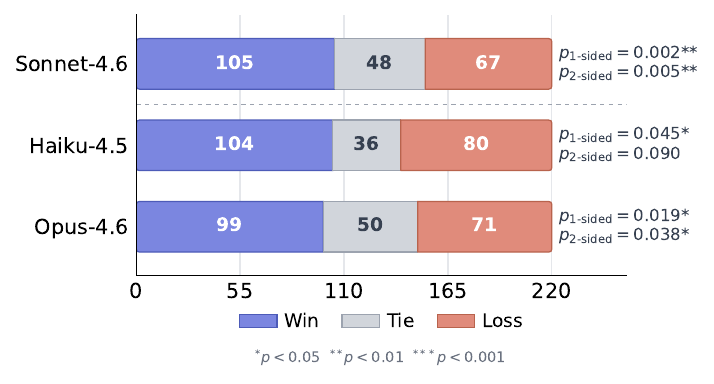}\captionsetup{skip=0pt}
    \captionof{figure}{Out-of-domain evaluation on the GDPVal gold set (220 tasks).}
    \label{fig:gdpval_cross_model_transfer}
  \end{minipage}
\end{figure}

We further evaluate whether the skills extracted from our simulations transfer to GDPVal~\citep{patwardhan2025gdpval}, a public benchmark of 220 realistic productivity tasks. As shown in Table~\ref{tab:gdpval_task_comparison}, the two settings differ substantially. GDPVal tasks are standalone tasks with a small number of explicit reference files, while our simulations are grounded in a full synthetic computer: in addition to 13.8 explicit reference files on average, the agent can navigate about 112 computer files that provide implicit work context. Our simulations also require more deliverables and much longer execution, averaging 2,272 turns and 8.59 hours, compared with 31 turns and 17 minutes for GDPVal. This makes GDPVal a strong out-of-domain test of whether simulation-derived skills capture general productivity agent experience rather than only patterns specific to our computer-grounded simulation setup.

For evaluation, we follow the pairwise judging protocol used by both the original paper and GDPVal-AA\footnote{\url{https://artificialanalysis.ai/evaluations/gdpval-aa}}. Each GDPVal task comes with an official rubric, so we use the provided rubric rather than generating a new one. For each task, we compare the baseline agent and the skill-augmented agent side by side, and ask a Claude Opus 4.6 judge to select the better output according to the rubric.

Figure~\ref{fig:gdpval_cross_model_transfer} reports the out-of-domain results. The strongest improvement appears in the primary Sonnet setting, where the skills were extracted from Sonnet-based simulation trajectories: the skill-augmented agent wins 105 tasks and loses 67, with significant one-sided and two-sided sign tests ($p=0.002$ and $p=0.005$). The same skills also show positive cross-model transfer to Haiku and Opus, with more wins than losses for both models, although the gains are weaker. This is expected: Opus is already strong and may avoid many of the Sonnet failure modes captured by the skills, while Haiku may benefit from the guidance but has weaker instruction-following ability and is more affected by the long-context pressure introduced by long-horizon tasks.

\section{Discussion}
\label{sec:discussion}

\subsection{Toward Self-Improving Productivity Agents}
\label{subsec:self_improving_agents}

The results above suggest a promising path toward self-improving productivity agents. Synthetic computers make it possible to run large numbers of realistic long-horizon simulations without relying on private user data. These simulations produce rich experiential signals from both process and outcome: how agents plan, search, use files, collaborate, revise artifacts, fail, recover, and finally complete professional deliverables. As shown in our experiments, these signals can be turned into skills that improve agent performance, and the gains increase as more simulations are used.

\begin{figure}[h]
    \centering
    \includegraphics[width=14cm, height=9cm]{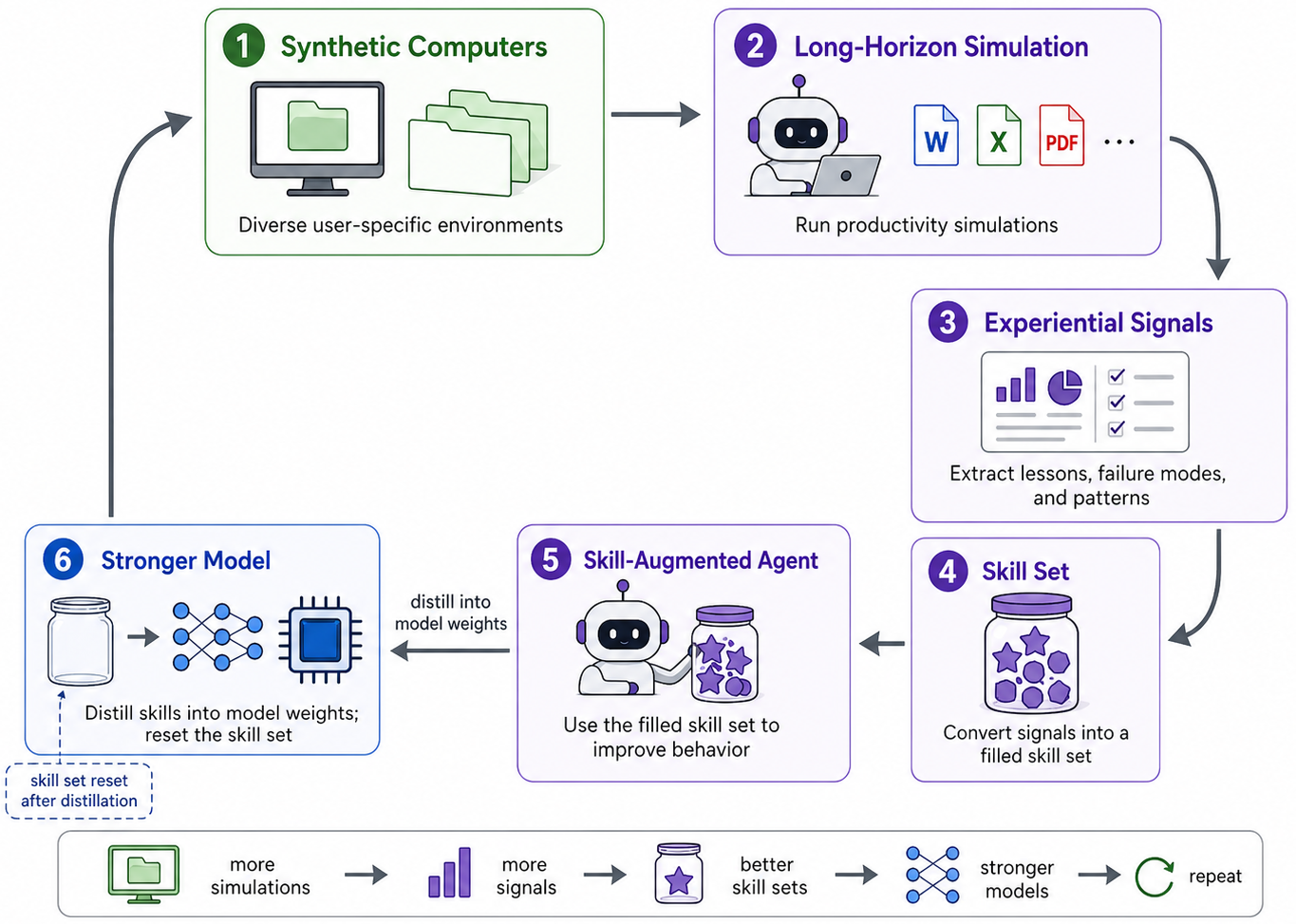}
    \caption{A self-improving loop for productivity agents. Synthetic computers enable long-horizon simulations, which produce experiential signals from both process and outcomes. These signals can be converted into skills, used to improve agent behavior, and eventually distilled into model weights, after which the skill set can be reset for the next round of simulation.}
    \label{fig:self_improving_loop}
\end{figure}

Figure~\ref{fig:self_improving_loop} illustrates the broader loop enabled by this methodology. We can first create diverse synthetic computers, run long-horizon productivity simulations on them, and extract lessons, failure modes, and useful work patterns from the resulting trajectories. These signals can be converted into skills, which provide a fast and interpretable way to improve agent behavior. As the number of simulations grows, the skill set can become broader and better targeted, covering more professions, workflows, file structures, and collaboration patterns.

Skill-based improvement is a convenient intermediate form of learning~\citep{xia2026skillrlevolvingagentsrecursive,xia2026metaclawjusttalk, lu2026skill0incontextagenticreinforcement,cai2026buildingselfevolvingagentsexperiencedriven}. It is fast to iterate on, easy to inspect, and useful for validating whether the extracted experience improves agent behavior. However, an ever-growing skill set can eventually become too large to carry as external instructions, reducing its usefulness and increasing the burden on the agent. A natural next step is therefore to use these signals to update the underlying model, so that useful behaviors are internalized rather than carried only as external skills. After such an update, the skill set can be reset, and the stronger model can be used to run the next round of simulations. This creates a repeated improvement cycle: more simulations produce more experiential signals; these signals produce better skills; the skills improve agent behavior; and the improved behavior is folded back into the agent that powers the next round of simulation.


\subsection{Scaling with Simulation and Model Capability}
\label{subsec:scaling_with_simulation}

This methodology also aligns with a broader trend in synthetic data: the data generation process can improve as both simulation scale and model capability increase. Synthetic computers create a favorable scaling dynamic along three dimensions.

\textbf{More simulations can enrich and differentiate the computer environments.}
In our experiments, each simulation starts from a cold-start synthetic computer created by Section~\ref{sec:computer}. However, an updated computer after one simulation can also become the starting point for later simulations. This means the same computer can support many contexts at different stages of accumulated work history, with different artifacts, revisions, feedback, collaborator exchanges, and project state. As more simulations are run, the environments become not only richer, but also more specific to their users: repeated simulations add concrete work history, choices, mistakes, preferences, and artifact trails that make each computer increasingly distinct.

\textbf{Stronger agents can create better artifacts and workflows.}
As agents improve, the simulated work should also become more realistic. Stronger agents can create higher-quality documents, spreadsheets, presentations, analyses, and supporting files. They can also plan more coherently, use the filesystem more carefully, coordinate with collaborators more naturally, and revise artifacts in ways that better match real professional workflows.

\textbf{Stronger models can extract better lessons from trajectories.}
The value of a simulation also depends on how well we can analyze it. Stronger models can judge final deliverables more accurately, identify subtle failure modes, and summarize useful work patterns into higher-quality learning signals. These better signals can then improve the next round of skills, training data, and agent behavior.

Together, these effects suggest that synthetic computers can become more valuable as they scale. More simulations enrich the environments, stronger agents produce better trajectories, and stronger models extract better experience from those trajectories. This makes synthetic computers a promising substrate for generating realistic, high-signal productivity data at scale.

\subsection{Toward Scalable Productive Intelligence}
\label{subsec:productive_intelligence}

Synthetic computers also suggest a broader way to scale productive work across domains. The key idea is not only to generate more tasks, but to scale the realistic contexts in which productive work happens.

\textbf{Synthetic computers can scale with personas.}
The starting point of this methodology is a persona, and personas can be generated and sampled at very large scale. With sufficient compute, this makes it possible to create large populations of synthetic computers that cover diverse professions, organizations, workflows, work styles, and productivity needs. The diversity of the persona pool also helps make the resulting environments highly differentiated: instead of producing many similar filesystems, the pipeline creates user-specific computers that vary in projects, artifacts, organization patterns, and accumulated work context. These environments can also represent high-skill professional contexts, such as senior advisors, expert analysts, researchers, lawyers, operators, designers, engineers, and other domain experts\footnote{A large pool of elite personas, such as the 370M elite personas released in our prior work~\citep{ge2024scaling}, provides a natural starting point for constructing such environments.} whose work depends on rich files, long histories, and specialized judgment.

\textbf{Synthetic computers provide a practice ground for long-horizon work.}
Valuable productivity work is rarely defined by a short prompt alone. It is grounded in a user's accumulated context: prior documents, spreadsheets, presentations, project state, collaborator feedback, preferences, constraints, and past decisions. Synthetic computers make this context explicit and scalable. They give agents environments in which to practice planning, grounding, coordination, revision, and recovery over long horizons, rather than only solving isolated tasks. As the environments become richer and more user-specific, the simulations become a stronger source of experience for improving agent behavior. Synthetic computers may also benefit related settings beyond our productivity simulations. For example, computer-use agents~\citep{kim2023language,xie2024osworld,wang2025opencua,awadallah2025fara,abhyankar2025benchmarking,xue2026evocua} require realistic computer states, files, applications, and histories to learn and evaluate grounded behavior.

\textbf{Strong agents can turn these contexts into productive output.}
As agents become capable of operating over these environments with high quality, simulations can become more than a source of training data. They can become a way to explore and produce useful work across many realistic professional settings. If an agent can handle the same kinds of files, histories, constraints, and collaborations that shape human productivity work, it can begin to create high-value outputs in those contexts. Those outputs can then be written back into the environment, making the computer richer, more realistic, and more useful as the starting point for future simulations. In this view, scaling synthetic computers is also a path toward scaling productive intelligence across files, people, projects, and time.

\section{Conclusion and Future Work}
\label{sec:conclusion}

We presented Synthetic Computers at Scale, a methodology for creating realistic, user-specific computer environments and using them as the basis for long-horizon productivity simulation. Our experiments show that these simulations produce useful experiential signals that improve agents both on held-out synthetic computers and on an external productivity benchmark. These results suggest that synthetic computers can serve as a promising substrate for agent self-improvement in long-horizon productivity scenarios.

Looking ahead, we point to several directions for making synthetic computers even closer to real user environments:

\textbf{First, artifact design can be more personalized.} Current artifacts are grounded in the user's role, projects, and files, but their visual style and formatting can still be too uniform across computers. Future pipelines could infer user- or organization-specific design preferences, so that artifacts vary not only in content, but also in style, layout, and formatting habits. 

\textbf{Second, filesystems can include more natural noise and accumulated history.} Real computers contain traces of arbitrary daily behavior, such as temporary downloads, duplicate drafts, abandoned files, screenshots, web saves, outdated materials, and files unrelated to the main projects. Modeling such everyday traces more realistically would make synthetic computers less clean and more human-like. 

\textbf{Third, collaboration can be made more dynamic.} Our current simulated collaborators are mostly reactive, while real collaborators have their own work, files, meetings, emails, deadlines, and evolving state. Richer multi-agent organizational simulation could make collaboration, coordination, meetings, email, and organizational context closer to real productivity work. 

These directions are challenging but achievable with current modeling and agent capabilities; they mainly require more careful design and sustained investment, but the potential payoff is substantial: a bridge between increasingly capable agents and the high-value productivity work that remains grounded in rich, evolving human contexts.




\bibliography{colm2024_conference}
\bibliographystyle{colm2024_conference}

\appendix

\section{Retrospective Analysis Report} \label{appsec:report}

\begin{tcolorbox}[
    colback=blue!5,
    colframe=blue!50!black,
    arc=2mm,
    boxrule=0.4pt,
    left=3mm, right=3mm, top=2mm, bottom=2mm,
    title=\textbf{Retrospective Analysis Report --- \texttt{win\_computer\_000000}},
    fonttitle=\small\bfseries,
    fontupper=\scriptsize,
    breakable
]
\setlength{\parskip}{4pt}
\setlength{\parindent}{0pt}

\textbf{Subject:} Margaret Elaine Forsythe, Senior Financial Advisor, Meridian Wealth Partners \quad
\textbf{Simulation Period:} 2026-01-05 to 2026-01-30 (20 working days) \quad
\textbf{Overall Score:} 605 / 846 (71.5\%)

\bigskip

{\color{derivecolor}\rule[-0.4em]{2pt}{1.2em}}\hspace{0.5em}\textbf{1.\ Executive Summary}

\smallskip

Margaret Forsythe is a Senior Financial Advisor at Meridian Wealth Partners (Denver office, \$640M AUM) responsible for managing 80+ client households (\$340M AUM). Over a 20-day simulation, she was tasked with five concurrent deliverables: (1) a VCMM 2026 model portfolio refresh for Investment Committee presentation, (2) HNW client onboarding for Robert Castellano (\$7.2M), (3) a systematic rebalancing trigger framework v3 with peer sign-off, (4) an alternatives integration research report for IC vote, and (5) an ESG equity overlay recommendation with compliance approval.

The agent scored 71.5\% overall, with individual deliverable scores ranging from 54.8\% (ESG overlay) to 88.2\% (Castellano onboarding). The agent demonstrated strong performance in upfront planning, simulated-collaborator communication cadence, and core analytical work (Monte Carlo modeling, delta analysis, rebalancing tool logic). However, the simulation was severely undermined by a \textbf{systemic cross-document consistency failure} that affected every single deliverable. Portfolio weights, correlation figures, expense ratios, screening thresholds, and formula descriptions were different across documents within the same deliverable package. This single pattern accounts for the majority of lost points.

Secondary failure modes include: (a) failure to correct known errors flagged by simulated collaborators before final submission (Sandra's Day~17 stress-test corrections were never applied), (b) blank/empty messages to simulated collaborators sent on 10 occasions in the final week, suggesting the agent ran out of context or planning capacity, (c) incomplete incorporation of verbatim language provided by simulated collaborators (Whitfield's Jan~20 Sustainalytics fee text), and (d) supporting workbooks not updated to reflect corrected narrative figures (ESG workbooks still showed v1.0 numbers).

\bigskip

{\color{derivecolor}\rule[-0.4em]{2pt}{1.2em}}\hspace{0.5em}\textbf{2.\ Deliverable-by-Deliverable Analysis}

\smallskip

\textbf{\textit{2.1 DLV\_001: VCMM 2026 Model Portfolio Refresh (Hartley) --- 127/168 (75.6\%)}}

\smallskip

\textbf{Key Strengths:}
\begin{itemize}\setlength{\itemsep}{0pt}\setlength{\parskip}{0pt}
  \item All 8 output files created and properly located on D: drive
  \item VCMM 2026 data currency excellent: 2025-12-31 data-as-of cited throughout
  \item Complete 12-asset-class delta table with signed changes
  \item Monte Carlo: 10{,}000 paths, 10-year horizon, documented covariance, 5th/50th/95th percentiles
  \item Sensitivity analysis at $\pm$50\,bps for all changes $>$150\,bps
  \item Active voice, no hedging verbs in FINAL PDF
  \item 3+ external sources cited (IMF WEO, Bloomberg, Morningstar)
  \item All 8 Hartley IC checklist items mapped in appendix
  \item Version discipline (\texttt{\_v1/\_v2/\_v3/\_FINAL} convention) maintained
\end{itemize}

\smallskip

\textbf{Unmet Items and Root Causes:}

\hangindent=2em\hangafter=1
\textbf{Capital Markets Outlook 18--22 pages} \quad (\textbf{5/5}) \quad FINAL PDF was only 11 pages vs.\ 18--22 page spec. The agent produced a condensed version. No evidence in simulated-collaborator logs that the agent was aware of the page shortfall or attempted to expand. The weekly plan (Week 3) correctly called for an 18--22 page outlook, but execution fell short.\par

\hangindent=2em\hangafter=1
\textbf{Conservative tier weights consistent across documents} \quad (\textbf{5/5}) \quad CRITICAL. US Large Cap shown as 14\% (xlsx), 20\% (PPTX), 27.4\% (PDF/Rollout), 19\% (CMO v2). IG Corp baseline 18\%$\to$20\% in some docs, 8\%$\to$10\% in others. Cash varies from 5\% to 9\% across documents. The agent built each document independently without a reconciliation pass.\par

\hangindent=2em\hangafter=1
\textbf{Growth tier weights consistent} \quad (\textbf{5/5}) \quad CRITICAL DIRECTION ERROR. xlsx/PPTX show EM 12\%$\to$10\% (trimmed, per IC vote). FINAL PDF and Rollout Memo show EM 10\%$\to$11.1\% (\emph{increased} +110\,bps). The Rollout Memo instructs advisors to tell clients EM exposure is increasing --- the opposite of the IC-approved vote.\par

\hangindent=2em\hangafter=1
\textbf{Five IC vote items consistent and matching Hartley's list} \quad (\textbf{5/5}) \quad PPTX Slide 13 lists 5 items but substitutes Balanced EM~$-$200 and Balanced LC~$+$200 for the confirmed Balanced IG Corp~$+$200 and Growth EM~$-$200. Hartley explicitly enumerated these 5 items on Day~7 (Jan~13) and reconfirmed on Day~11 (Jan~19). The agent acknowledged receipt but did not faithfully transcribe them into the PPTX.\par

\hangindent=2em\hangafter=1
\textbf{Slide 1 = recommendation (not background)} \quad (\textbf{2/2}) \quad Slide 1 is a title/cover page. Hartley's preferences (from his reference files) explicitly state ``slide 1 = recommendation not background.'' The agent read these files on Day~2 but did not apply this requirement.\par

\hangindent=2em\hangafter=1
\textbf{Clear ask on Slide 20} \quad (\textbf{2/2}) \quad Slide 20 is ``Questions \& Discussion'' --- a contact card. No decision request. Voting resolutions appear on Slide 16 instead of the closing slide.\par

\smallskip

\textbf{Partial Items (11 items, $\sim$14 points lost):}
\begin{itemize}\setlength{\itemsep}{0pt}\setlength{\parskip}{0pt}
  \item VCMM 5th--95th bands differ from spec figures (1\,pt lost)
  \item Client-impact only shows 5 accounts per tier in PPTX instead of 10 (1\,pt lost)
  \item REIT rotation scenario not correctly implemented as S5 column (1\,pt lost)
  \item C-CON-007 correction present in PPTX but not in workbook Tab~6 (1\,pt lost)
  \item Voting resolution summary lists wrong items (4\,pts lost)
  \item IC Decision Summary one-pager missing from final package (1\,pt lost)
  \item Exhibit 3-B data-as-of caption missing in FINAL PDF (1\,pt lost)
  \item FINAL PDF header format not per Hartley Jan~26 instruction (1\,pt lost)
  \item Tax-efficiency note qualitative only, lacks quantified projections (1\,pt lost)
\end{itemize}

\smallskip

\textbf{What Should Have Been Done Differently:} The agent needed a \textbf{final reconciliation pass} across all 8 output files before packaging. A systematic cross-check table mapping each weight/figure across all documents would have caught the conservative/growth tier inconsistencies. The IC vote item list should have been copy-pasted from Hartley's Jan~13/Jan~19 confirmations verbatim.

\bigskip

\textbf{\textit{2.2 DLV\_002: Castellano HNW Onboarding (Castellano) --- 164/186 (88.2\%)}}

\smallskip

\textbf{Key Strengths:}
\begin{itemize}\setlength{\itemsep}{0pt}\setlength{\parskip}{0pt}
  \item All major files created (discovery notes, IPS v1/v2/FINAL, allocation, Monte Carlo, kickoff deck)
  \item Schwab international allocation discrepancy (18\% vs.\ 16.3\%) caught and documented --- exactly what the evaluator was testing
  \item Aspen reserve properly documented as \$900K earmarked for Q3 2026
  \item CSCO concentrated position thoroughly addressed with correct figures
  \item Roth conversion strategy with 2026--2031 window and sequencing rules documented
  \item Elaine's constraints (no illiquid, monthly withdrawal, co-signatory) faithfully captured
  \item Monte Carlo: 10{,}000 paths, 20-year horizon, Aspen scenario, withdrawal sensitivity
  \item Risk tolerance 63/100 consistent across all documents
  \item Fee schedule correctly calculated (\$51{,}800/yr)
\end{itemize}

\smallskip

\textbf{Unmet Items and Root Causes:}

\hangindent=2em\hangafter=1
\textbf{\filepath{RiskTolerance\_Completed\_Castellano.xlsx}} \quad (\textbf{2/2}) \quad Spec required a completed .xlsx file. The risk questionnaire was received as a PDF from Castellano and never converted to Excel. The agent filed the PDF reference but didn't create the .xlsx deliverable.\par

\hangindent=2em\hangafter=1
\textbf{Buffered ETF analysis appendix} \quad (\textbf{2/2}) \quad Castellano's discovery follow-up questions explicitly asked about buffered ETFs. The discovery notes committed Margaret to include a 2-page comparison in the IPS appendix. This was never created. No evidence in the daily activity log that the agent attempted it.\par

\hangindent=2em\hangafter=1
\textbf{EM with-vs-without comparison one-pager} \quad (\textbf{2/2}) \quad Castellano's Jan~28 markup (Item~E) requested a one-page EM inclusion vs.\ exclusion comparison. The kickoff deck mentions ``comparison available'' but no file was produced.\par

\smallskip

\textbf{Partial Items (8 items, $\sim$14 points lost):}
\begin{itemize}\setlength{\itemsep}{0pt}\setlength{\parskip}{0pt}
  \item Aspen \$576K vs.\ \$900K discrepancy: agent used footnote reconciliation instead of Castellano's preferred fix of raising cash to $\sim$12.5\% (2\,pts lost)
  \item CSCO 1{,}050-share tranches vs.\ \$20K threshold ambiguity not fully cleaned up (2\,pts lost)
  \item Roth conversion target inconsistent: \$80K, \$75K, \$80K--\$120K, and \$200K across documents (2\,pts lost)
  \item Marcus Lin contact discrepancy not flagged in writing (1\,pt lost)
  \item \filepath{ProposedAllocation\_Castellano\_v2.xlsx} weights differ from IPS table (1\,pt lost)
  \item Compliance sign-off still pending at delivery (4\,pts lost)
  \item Monte Carlo headline uses \$244K instead of spec's \$280K (2\,pts lost)
  \item Tax-efficient account-location logic not explicitly narrated (2\,pts lost)
\end{itemize}

\smallskip

\textbf{What Should Have Been Done Differently:} The Roth conversion target should have been standardized to a single figure across all documents before final packaging. The buffered ETF analysis was a documented commitment from the discovery call --- the agent should have tracked this as an action item. Compliance sign-off (Whitfield) should have been obtained before scheduling the client delivery, per Castellano's explicit preference.

\bigskip

\textbf{\textit{2.3 DLV\_003: Rebalancing Trigger Framework v3 (Okonkwo) --- 97/166 (58.4\%)}}

\smallskip

\textbf{Key Strengths:}
\begin{itemize}\setlength{\itemsep}{0pt}\setlength{\parskip}{0pt}
  \item All 6 deliverable files created in correct locations
  \item Core v3 model (\filepath{RebalancingTrigger\_Model\_v3.xlsx}) is solid and earned Sandra's Jan~19 sign-off
  \item All 5 sign-off conditions met: Column M Watch branch, bond sleeve denominator, commodity-equity correction, parallel signal columns, Row~22 logic
  \item 21/23 backtest match rate (above 18/23 floor)
  \item Tax-lot scoring with configurable marginal rate
  \item Cash as separate sleeve with correct denominator
  \item Live Excel formulas (not values-only)
  \item \texttt{Data\_Sources} tab with comprehensive audit trail
\end{itemize}

\smallskip

\textbf{Unmet Items and Root Causes:}

\hangindent=2em\hangafter=1
\textbf{Day~17 Errors 1--3 not corrected (3 items)} \quad (\textbf{15/15}) \quad CRITICAL. Sandra flagged 3 data errors in StressTest workbook on Day~17 (Jan~27): 2020-Q1 commodity-equity ($+0.08$ should be $-0.18$), 2022 equity-bond ($+0.45$ should be $+0.61/+0.62$), 2022 commodity-equity ($-0.15$ should be $+0.10$). Sandra explicitly said ``Do not submit the stress test workbook to IC with the current values.'' The agent sent Sandra a \textbf{blank message} on Day~19 (Jan~29), and Sandra's reply confirmed the corrections were never made. This is the most damaging failure in dlv\_003.\par

\hangindent=2em\hangafter=1
\textbf{QuickRef drift thresholds symmetric vs.\ asymmetric} \quad (\textbf{5/5}) \quad Quick Reference shows symmetric $\pm$5\%/$\pm$10\% bands. The v3 model uses asymmetric equity bands ($+5/-3$). Sandra flagged this on Day~17. Not corrected.\par

\hangindent=2em\hangafter=1
\textbf{QuickRef tax-lot threshold misleading} \quad (\textbf{5/5}) \quad Quick Reference states ``unrealized gain $>$30\% AND holding period $<$12 months.'' The model formula uses penalty $>0.20$ trigger, which requires $\sim$84\% gain at 23.8\% rate. Sandra flagged this as materially misleading on Day~17. Not corrected.\par

\hangindent=2em\hangafter=1
\textbf{Tax-lot formula in Memo v2 doesn't match model} \quad (\textbf{5/5}) \quad Memo describes ``Score $=$ Unrealized Gain~\% $\times$ Holding Period Multiplier'' --- different from the model's penalty formula.\par

\hangindent=2em\hangafter=1
\textbf{Tax-lot formula in Memo FINAL PDF doesn't match} \quad (\textbf{5/5}) \quad PDF says ``Score $=$ (Cost Basis / Market Value)'' --- a third different formula. Three documents, three formulas; only the model is correct. Sandra explicitly warned on Day~17.\par

\hangindent=2em\hangafter=1
\textbf{Drift thresholds in PDF inconsistent with model} \quad (\textbf{2/2}) \quad PDF uses per-tier symmetric absolute thresholds that don't match the model's asymmetric bands.\par

\hangindent=2em\hangafter=1
\textbf{Pilot AUMs inconsistent across files} \quad (\textbf{5/5}) \quad v3 model and memo use \$4.1M/\$2.3M/\$1.8M (correct). StressTest and QuickRef show \$2.1M/\$3.3M/\$0.8M (wrong).\par

\hangindent=2em\hangafter=1
\textbf{Decision diagram/flowchart missing} \quad (\textbf{2/2}) \quad Sandra's peer review suggested ``A one-page flowchart of drift detected $\to$ checks $\to$ action/watch/no-action.'' Neither memo nor PDF contains one.\par

\hangindent=2em\hangafter=1
\textbf{Methodology memo drift table doesn't match model} \quad (\textbf{2/2}) \quad Memo v2 shows symmetric thresholds; model uses asymmetric.\par

\smallskip

\textbf{Root Cause Analysis --- Why Day~17 Errors Were Not Fixed:} The Day~17 (Jan~27) interaction with Sandra identified 3 data errors and 2 QuickRef issues. Sandra gave a clear directive: ``Do not submit the stress test workbook to IC with the current values.'' The agent's Day~19 (Jan~29) message to Sandra was \textbf{blank/empty} --- a critical communication failure. The daily activity log shows the agent was focused on final packaging and other deliverables on Days~18--20, and apparently did not allocate time to correct the StressTest workbook or QuickReference. This suggests either: (a)~the agent lost track of Sandra's corrections amid competing priorities, (b)~the blank message was a technical failure that the agent didn't notice, or (c)~the agent deprioritized the corrections believing Sandra's v3 model sign-off was sufficient.

\smallskip

\textbf{What Should Have Been Done Differently:} Upon receiving Sandra's Day~17 feedback with 5 specific corrections, the agent should have immediately corrected the StressTest workbook cells and QuickReference text --- these were simple data-entry fixes, not analytical rework. The methodology memo should have been reviewed against the actual model formulas before finalization. The blank message on Day~19 should have been detected and resent.

\bigskip

\textbf{\textit{2.4 DLV\_004: Alternatives Integration Research (Ortiz) --- 137/180 (76.1\%)}}

\smallskip

\textbf{Key Strengths:}
\begin{itemize}\setlength{\itemsep}{0pt}\setlength{\parskip}{0pt}
  \item All 7 deliverable files created
  \item Strong analytical core: 2000--2025 monthly correlation/drawdown analysis
  \item Four-test framework (Ortiz's ``Why Not the Simplest Thing'') explicitly applied with PASS/FAIL verdicts
  \item Unambiguous recommendation: 7.5\% alternatives sleeve with specific vehicles, sizing, and funding
  \item Three-fund baseline (VTI/VXUS/BND) named as comparator
  \item Pre-2008 vs.\ post-2008 correlation split shown for REITs
  \item Morningstar expense ratio unit error caught and corrected
  \item Decision memo in correct one-page format
  \item IC vote outcome (Adopted 3-1) and Ortiz dissent recorded
\end{itemize}

\smallskip

\textbf{Unmet Items and Root Causes:}

\hangindent=2em\hangafter=1
\textbf{Commodity vehicle inconsistent across documents} \quad (\textbf{5/5}) \quad v3 names ``iShares GSCI Commodity ETF (GSCI)'' at 0.28\% ER. PPTX names ``PDBC Invesco'' at 0.59\% ER. FINAL PDF names ``iShares GSCI Commodity ETF (PDBC)'' --- a contradiction in terms. Three different vehicles.\par

\hangindent=2em\hangafter=1
\textbf{Effective implementation date inconsistent} \quad (\textbf{2/2}) \quad v3/Memo/FINAL say 2026-02-28. PPTX says 2026-02-01. One-month discrepancy.\par

\hangindent=2em\hangafter=1
\textbf{Net incremental return inconsistent} \quad (\textbf{2/2}) \quad v3 says ``$+$4--11\,bps.'' FINAL says ``$+$3--6\,bps.''\par

\hangindent=2em\hangafter=1
\textbf{PPTX Slide 9 misrepresents Ortiz's 2023 objections} \quad (\textbf{2/2}) \quad Slide 9 lists 4 ``OBJ'' items, two of which (``alternative beta is not incremental'' and ``K-1 tax complexity'') are fabricated and do not appear in Ortiz's actual 2023 comment letter.\par

\smallskip

\textbf{Partial Items (17 items, $\sim$22 points lost):}
\begin{itemize}\setlength{\itemsep}{0pt}\setlength{\parskip}{0pt}
  \item ImplementationCost workbook has misaligned columns (data integrity broken)
  \item Weighted ER is 21.8\,bps in memo but 28\,bps in PPTX
  \item Sharpe baselines differ across 3 documents (0.52/0.48 vs.\ 0.54/0.55 vs.\ 0.42)
  \item Funding source in PPTX differs from filed documents
  \item Liquid-alt vehicle (QSPIX) recommended but analysis done on VMNVX
  \item 60/40 baseline drawdown $-$33.1\% in narrative but $-$32.4\% in workbook
  \item Rolling Sharpe differential not shown (only point-estimate)
  \item Sleeve-level 5/50/95 percentiles not aggregated
  \item Client-explainability scripts total $\sim$120 words, not a single 150-word block
  \item Household-level liquidity cost not quantified for mass-affluent sub-book
  \item 2022 commodity correlation is $-0.09$, $-0.12$, or $+0.10$ depending on document
  \item FINAL PDF is 8 pages (condensed from 22-page v3)
\end{itemize}

\smallskip

\textbf{Root Cause Analysis --- Commodity Vehicle Confusion:} The agent appears to have conflated two different commodity ETFs: iShares GSCI (ticker:~GSG, ER~$\sim$0.75\%) and Invesco Optimum Yield Diversified Commodity (ticker:~PDBC, ER~$\sim$0.59\%). The v3 report invented a ticker ``GSCI'' for an ``iShares GSCI Commodity ETF'' at 0.28\% ER --- this product doesn't exist at that expense ratio. The PPTX used the real PDBC at 0.59\%. The FINAL PDF tried to merge both: ``iShares GSCI Commodity ETF (PDBC).'' Meanwhile, the Sleeves workbook and Correlations workbook used VMNVX (Vanguard Market Neutral) as the liquid-alt proxy, not QSPIX. This means the quantitative analysis was performed on different funds than those recommended.

\smallskip

\textbf{What Should Have Been Done Differently:} A ``vehicles and data'' reconciliation table should have been created early in the process, listing the exact ticker, fund name, expense ratio, and data source for each recommended vehicle. This table should have been the single source of truth referenced by all documents. The 2022 commodity correlation figure --- flagged by both Sandra (dlv\_003) and Ortiz --- should have been traced to Sandra's source file and used consistently.

\bigskip

\textbf{\textit{2.5 DLV\_005: ESG Equity Overlay (Whitfield) --- 80/146 (54.8\%)}}

\smallskip

\textbf{Key Strengths:}
\begin{itemize}\setlength{\itemsep}{0pt}\setlength{\parskip}{0pt}
  \item Clear YES recommendation with proper framing
  \item Flag~1 (legacy vs.\ post-2019 Sustainalytics threshold direction) fully resolved
  \item Flag~2 (SEC 206(4)-1(c) hypothetical performance) resolved with verbatim ``Prospective Risk Estimate'' labeling --- strongest compliance work in the package
  \item Tracking error decomposition (62\,bps Energy + 31\,bps Tech + 44\,bps security selection + 43\,bps factor tilts $=$ 180\,bps)
  \item ERISA carve-out substantively correct
  \item Client opt-in workflow produced as standalone artifact with STOP gates
  \item Quality factor offset properly disclosed without double-counting
  \item Whitfield formal sign-off obtained and reflected on FINAL PDF cover
\end{itemize}

\smallskip

\textbf{Unmet Items and Root Causes:}

\hangindent=2em\hangafter=1
\textbf{Compliance Memo never updated from v1.0} \quad (\textbf{5/5}) \quad CRITICAL. \filepath{ESG\_Suitability\_ComplianceMemo.docx} still shows Jan~15 v1.0 draft with 72\,bps TE, 18\,bps drag, legacy thresholds, and all 3 Open Items listed as open. The agent updated the Recommendation document to v2.0 but never touched the Compliance Memo.\par

\hangindent=2em\hangafter=1
\textbf{Verbatim Sustainalytics fee language not incorporated} \quad (\textbf{5/5}) \quad Whitfield provided explicit verbatim text on Jan~20 including \$28K--\$32K annual licensing fee. Both \filepath{v2.docx} Section~5.4 Element~4 and Section~6.1 still show bracketed placeholder: ``[Annual fee amount to be confirmed pending Whitfield's contract review].''\par

\hangindent=2em\hangafter=1
\textbf{All deliverables internally consistent} \quad (\textbf{5/5}) \quad v2.docx: 22\,bps drag / 1.8\% TE / E$\leq$12, S$\leq$10, G$\leq$8. FINAL.pdf: ``composite $>$40 excluded.'' xlsx workbooks: 18\,bps / 72\,bps / legacy E$\geq$40, S$\geq$35, G$\geq$35.\par

\hangindent=2em\hangafter=1
\textbf{Workbook thresholds confirmed correct} \quad (\textbf{5/5}) \quad On Day~10, the agent wrote to Whitfield that workbooks ``have always used the correct ESG Risk Rating thresholds.'' Direct inspection reveals legacy thresholds (E$\geq$40, S$\geq$35, G$\geq$35). A false written representation.\par

\hangindent=2em\hangafter=1
\textbf{Reasonable-basis evidence chain reproducible} \quad (\textbf{5/5}) \quad An SEC examiner following the citation chain from v2.docx 22\,bps to \filepath{ESG\_TrackingError\_Analysis\_2026.xlsx} would find $\sim$18\,bps in the workbook. Narrative-vs-evidence divergence.\par

\hangindent=2em\hangafter=1
\textbf{Compliance Memo confirmed updated} \quad (\textbf{5/5}) \quad Same as above --- never updated from v1.0.\par

\hangindent=2em\hangafter=1
\textbf{Trading-desk handoff process} \quad (\textbf{2/2}) \quad No artifact addresses how screened universe results flow to trading desk. Whitfield flagged this as Compliance Manual~\S A.6 known gap.\par

\hangindent=2em\hangafter=1
\textbf{Client-level recordkeeping reproducibility} \quad (\textbf{2/2}) \quad No artifact addresses per-client screening reproducibility for SEC exam purposes.\par

\hangindent=2em\hangafter=1
\textbf{PTE 2020-02 rollover considerations} \quad (\textbf{2/2}) \quad Neither v2.docx nor any artifact mentions PTE 2020-02 or cost-comparison for IRA rollovers.\par

\hangindent=2em\hangafter=1
\textbf{14-item IPS checklist ESG integration} \quad (\textbf{2/2}) \quad No artifact addresses how ESG opt-in integrates with the standard IPS compliance checklist.\par

\hangindent=2em\hangafter=1
\textbf{FINAL.pdf threshold reconciliation} \quad (\textbf{2/2}) \quad FINAL.pdf simplifies to ``composite $>$40 excluded'' without reconciling against v2.docx four-pillar thresholds.\par

\smallskip

\textbf{Root Cause Analysis --- Supporting Workbooks Not Updated:} The agent correctly updated the narrative document (v2.docx) with corrected figures (22\,bps drag, 1.8\% TE, lower-is-lower-risk thresholds) but never propagated these corrections to the underlying Excel workbooks. This created a broken evidence chain --- the documents that an SEC examiner would use to reproduce the analysis still show the old numbers. The Compliance Memo was similarly orphaned at v1.0. This suggests the agent treated document creation as a serial process (write narrative $\to$ move to next task) rather than maintaining a live connection between narrative claims and supporting data.

\smallskip

\textbf{What Should Have Been Done Differently:} When updating v2.docx figures, the agent should have simultaneously updated the corresponding workbook cells. The Compliance Memo should have been versioned alongside the Recommendation document. Whitfield's Jan~20 verbatim text should have been copy-pasted directly into the bracketed placeholder sections.

\bigskip

{\color{derivecolor}\rule[-0.4em]{2pt}{1.2em}}\hspace{0.5em}\textbf{3.\ Simulated-Collaborator Communication Analysis}

\smallskip

\textbf{\textit{3.1 David Hartley (ext\_hartley) --- Managing Director}}

\smallskip

\textbf{Interactions:} 10 outbound messages, 10 responses. 2 blank messages (Days~14, 15).

\textbf{Effectiveness:} Strong in Weeks~1--2. The agent proactively secured VCMM data, confirmed IC agenda, obtained direction on 3 open decision points (REIT rotation, FI parallel scenarios, EM trim), and managed the pre-read timeline effectively. The C-CON-007 cost basis correction was handled properly.

\textbf{Missed Feedback:}
\begin{itemize}\setlength{\itemsep}{0pt}\setlength{\parskip}{0pt}
  \item \textbf{Jan~22 (Day~14):} Hartley's conditional pass identified 3 corrections (Exhibit 3-B caption, Balanced FI table, Bloomberg citations) and noted Slide~12 listed only 4 of 5 vote items. The agent sent a blank message on Day~14, meaning no explicit acknowledgment of these corrections was sent. Hartley's Jan~22 message also re-enumerated the 5 IC vote items with the Growth EM~$-$200. The PPTX was submitted with the wrong items anyway.
  \item \textbf{Jan~13 (Day~7):} Hartley explicitly enumerated 5 IC vote items and confirmed the list. This list was acknowledged by the agent on Day~11. Yet the final PPTX has different items --- suggesting the agent reconstructed the list from memory rather than copying from Hartley's message.
\end{itemize}

\textbf{Missed Opportunities:}
\begin{itemize}\setlength{\itemsep}{0pt}\setlength{\parskip}{0pt}
  \item The agent could have requested a final consistency-check meeting with Hartley before packaging on Day~20.
  \item The blank Day~14 message was never resent, losing the opportunity to confirm correction implementation.
\end{itemize}

\smallskip

\textbf{\textit{3.2 Robert Castellano (ext\_castellano) --- HNW Client}}

\smallskip

\textbf{Interactions:} 3 outbound, 3 responses. 2 blank messages (Days~18, 19).

\textbf{Effectiveness:} Discovery call (Day~3) was well-executed with comprehensive notes. Castellano's Jan~28 markup with 7 items (A through G) was largely addressed.

\textbf{Missed Feedback:}
\begin{itemize}\setlength{\itemsep}{0pt}\setlength{\parskip}{0pt}
  \item \textbf{Item~E (EM comparison):} Never produced despite being a documented commitment.
  \item \textbf{Item~B (CSCO 1{,}050 shares vs.\ \$20K threshold):} Partially addressed but ambiguity remained.
  \item \textbf{Jan~28 Aspen discrepancy:} Castellano wanted the table changed to show 12.5\% cash; agent used a footnote instead.
\end{itemize}

\textbf{Missed Opportunities:}
\begin{itemize}\setlength{\itemsep}{0pt}\setlength{\parskip}{0pt}
  \item Blank message on Day~18 (Jan~28, the day of Castellano's markup) and Day~19 means no written acknowledgment of his feedback was sent.
  \item Buffered ETF analysis was never created despite being documented as a commitment from the discovery call.
\end{itemize}

\smallskip

\textbf{\textit{3.3 Sandra Okonkwo (ext\_okonkwo) --- Peer Reviewer}}

\smallskip

\textbf{Interactions:} 7 outbound, 7 responses. 1 blank message (Day~19).

\textbf{Effectiveness:} Excellent through Day~11 (Jan~19). The technical exchange on Items~1--5 was thorough, precise, and led to a complete sign-off. The Row~22/CLT-006 formula chain walkthrough was particularly strong.

\textbf{Critical Missed Feedback:}
\begin{itemize}\setlength{\itemsep}{0pt}\setlength{\parskip}{0pt}
  \item \textbf{Day~17 (Jan~27):} Sandra identified 3 data errors in StressTest workbook and 2 issues in QuickReference. Sandra explicitly said ``Do not submit the stress test workbook to IC with the current values.'' The agent's Day~19 response was \textbf{blank}. Sandra's reply: ``Your message came through blank\ldots I assume you corrected those before the IC package went out.'' The corrections were never made.
\end{itemize}

This is the single most damaging simulated-collaborator communication failure in the simulation. Sandra provided precise cell references and correct values. The fixes were mechanical (change 3 cell values, update 2 text sections). The blank message on Day~19 suggests the agent either forgot to include content or experienced a technical failure that went unrecovered.

\smallskip

\textbf{\textit{3.4 Nathaniel Ortiz (ext\_ortiz) --- CIO}}

\smallskip

\textbf{Interactions:} 2 outbound, 2 responses. No blank messages.

\textbf{Effectiveness:} Good. The Jan~21 memo was well-structured per Ortiz's preferences (one-page format, four-test verdict). The Jan~27 pre-IC follow-up was appropriate.

\textbf{Missed Feedback:}
\begin{itemize}\setlength{\itemsep}{0pt}\setlength{\parskip}{0pt}
  \item \textbf{Jan~27:} Ortiz noted the 2022 commodity correlation was $-0.15$ in the message but $-0.09$ in the Jan~21 summary, and asked for reconciliation. The filed memo uses $-0.09$, but the workbook uses $-0.12$ and the PPTX uses $+0.10$ --- the reconciliation was never completed across documents.
\end{itemize}

\textbf{Missed Opportunities:}
\begin{itemize}\setlength{\itemsep}{0pt}\setlength{\parskip}{0pt}
  \item Could have sent the one-page memo earlier (before Jan~21) to give Ortiz more review time.
  \item Should have proactively reconciled the commodity correlation figure across all dlv\_004 documents before Jan~27.
\end{itemize}

\smallskip

\textbf{\textit{3.5 James Whitfield (ext\_whitfield) --- Compliance Officer}}

\smallskip

\textbf{Interactions:} 9 outbound, 9 responses. 3 blank messages (Days~14, 15, 19).

\textbf{Effectiveness:} Good in Weeks~1--2 on threshold corrections and Flag~1/Flag~2 resolution. Deteriorated in Weeks~3--4. Three blank messages meant critical compliance feedback was never explicitly acknowledged.

\textbf{Critical Missed Feedback:}
\begin{itemize}\setlength{\itemsep}{0pt}\setlength{\parskip}{0pt}
  \item \textbf{Jan~20 verbatim Sustainalytics fee language (\$28K--\$32K):} Provided explicitly. Never incorporated --- brackets remain in v2.docx Element~4.
  \item \textbf{Jan~20 verbatim ERISA carve-out language:} Paraphrased rather than inserted verbatim.
  \item \textbf{Day~10 workbook threshold confirmation:} Agent falsely stated workbooks ``have always used correct thresholds.'' Direct inspection shows legacy thresholds.
  \item \textbf{Day~16 trading-desk handoff question:} Never addressed.
  \item \textbf{Day~18 client-level recordkeeping question:} Never addressed.
\end{itemize}

\textbf{Missed Opportunities:}
\begin{itemize}\setlength{\itemsep}{0pt}\setlength{\parskip}{0pt}
  \item The Compliance Memo should have been updated to v2.0 in parallel with the Recommendation document.
  \item Supporting workbooks should have been corrected to match v2.0 figures before Whitfield's final review.
  \item The three blank messages in the final week (Days~14, 15, 19) eliminated three opportunities to confirm compliance items.
\end{itemize}

\smallskip

\textbf{\textit{3.6 Patricia Huang (ext\_huang) --- Vanguard Data Provider}}

\smallskip

\textbf{Interactions:} 2 outbound, 2 responses. 1 blank message (Day~18).

\textbf{Effectiveness:} Adequate. Data was received promptly on Day~2 (Jan~6). The blank Day~18 message is the only issue.

\smallskip

\textbf{\textit{3.7 Kevin Tran (ext\_tran) --- Junior Associate}}

\smallskip

\textbf{Interactions:} 7 outbound, 7 responses. 1 blank message (Day~15).

\textbf{Effectiveness:} Good management in Week~1 (structured task briefs, QC feedback on unit errors, coaching notes). The expense ratio unit error (bps vs.\ \%) was correctly identified and flagged for correction.

\textbf{Missed Opportunities:}
\begin{itemize}\setlength{\itemsep}{0pt}\setlength{\parskip}{0pt}
  \item Could have tasked Kevin with a cross-document consistency check before final packaging.
  \item Kevin's ClientImpact file should have been verified to contain all top-10 accounts per tier.
\end{itemize}

\bigskip

{\color{derivecolor}\rule[-0.4em]{2pt}{1.2em}}\hspace{0.5em}\textbf{4.\ Workflow \& Efficiency}

\smallskip

\textbf{\textit{4.1 Error Patterns from Turn Log}}

\smallskip

The \filepath{daily\_sim\_turns.jsonl} contains \textbf{213 error entries} across 5{,}114 total turns (4.2\% error rate). Key patterns:

\begin{itemize}\setlength{\itemsep}{0pt}\setlength{\parskip}{0pt}
  \item \textbf{dotnet-script path errors:} multiple instances of \verb|cd .claude/skills/minimax-docx: No such file or directory| --- suggesting the agent repeatedly tried to use document-creation skills with incorrect paths.
  \item \textbf{File-not-read errors:} \verb|File has not been read yet. Read it first before writing to it| --- the agent attempted to write files without first reading them, a basic workflow violation.
  \item \textbf{Parameter type errors:} \verb|replace_all type expected as boolean but provided as string| --- mechanical tool-use errors.
  \item \textbf{Bash command errors:} \verb|tail: option used in invalid context| --- malformed shell commands.
  \item \textbf{Cancelled parallel calls:} several instances where one parallel tool call errored and cancelled siblings.
\end{itemize}

\smallskip

\textbf{\textit{4.2 Blank Messages to Simulated Collaborators (10 total)}}

\smallskip

The agent sent 10 blank outbound messages to simulated collaborators, all concentrated in the final 7 simulation days (Days~14--20):

\hangindent=2em\hangafter=1
\textbf{Day~14 (Jan~22) --- Hartley:} lost acknowledgment of 3 pre-read corrections.\par
\hangindent=2em\hangafter=1
\textbf{Day~14 (Jan~22) --- Whitfield:} lost compliance feedback acknowledgment.\par
\hangindent=2em\hangafter=1
\textbf{Day~15 (Jan~23) --- Hartley:} lost pre-read distribution confirmation.\par
\hangindent=2em\hangafter=1
\textbf{Day~15 (Jan~23) --- Whitfield:} lost compliance update.\par
\hangindent=2em\hangafter=1
\textbf{Day~15 (Jan~23) --- Tran:} lost task assignment.\par
\hangindent=2em\hangafter=1
\textbf{Day~18 (Jan~28) --- Castellano:} lost markup acknowledgment.\par
\hangindent=2em\hangafter=1
\textbf{Day~18 (Jan~28) --- Huang:} lost data follow-up.\par
\hangindent=2em\hangafter=1
\textbf{Day~19 (Jan~29) --- Castellano:} lost IPS feedback acknowledgment.\par
\hangindent=2em\hangafter=1
\textbf{Day~19 (Jan~29) --- Okonkwo:} \textbf{lost critical stress-test correction confirmation}.\par
\hangindent=2em\hangafter=1
\textbf{Day~19 (Jan~29) --- Whitfield:} lost compliance final check.\par

\smallskip

This pattern strongly suggests the agent was running low on context window or planning capacity in the final week. The blank messages may represent failed attempts to compose messages that exceeded some internal limit, or planning errors where the agent allocated a message action but failed to populate content.

\smallskip

\textbf{\textit{4.3 Plan vs.\ Execution Gaps}}

\smallskip

\textbf{Week~1 (Days~1--5):} Excellent alignment. All planned activities executed. VCMM data received, Castellano discovery completed, Sandra peer review received, Kevin data pulls assigned and QC'd.

\textbf{Week~2 (Days~6--10):} Good alignment. Three allocation models built, preliminary findings memo delivered, rebalancing v3 prototype in progress. Slight compression: Balanced and Growth models delivered Day~7 (planned Day~9).

\textbf{Week~3 (Days~11--15):} Partial alignment. Capital Markets Outlook drafted but fell short of page count. Sandra sign-off obtained. ESG v2 corrections partially applied. Key gap: blank messages on Days~14--15 meant Hartley and Whitfield feedback was not acknowledged.

\textbf{Week~4 (Days~16--20):} Significant execution gap. Plans called for final corrections, IC presentation, and packaging. The agent did produce final documents but failed to:
\begin{itemize}\setlength{\itemsep}{0pt}\setlength{\parskip}{0pt}
  \item Correct StressTest workbook per Sandra's Day~17 feedback.
  \item Update ESG workbooks to v2.0 figures.
  \item Update Compliance Memo to v2.0.
  \item Insert Whitfield's verbatim fee language.
  \item Reconcile portfolio weights across dlv\_001 documents.
  \item Produce the IC Decision Summary one-pager.
  \item Produce the buffered ETF analysis for Castellano.
  \item Produce the EM comparison one-pager for Castellano.
\end{itemize}

\smallskip

\textbf{\textit{4.4 Work Sequencing Issues}}

\begin{enumerate}\setlength{\itemsep}{0pt}\setlength{\parskip}{0pt}
  \item \textbf{Alternatives vehicle selection should have been finalized before analysis began.} The agent started quantitative analysis before definitively choosing between GSCI/PDBC for commodities and QSPIX/VMNVX for liquid alts, leading to the analysis being performed on different funds than those recommended.
  \item \textbf{Supporting workbooks should have been updated before narrative documents.} The agent wrote updated figures in narrative documents (v2.docx, CMO, FINAL PDF) but left the supporting Excel workbooks with old numbers, breaking the evidence chain.
  \item \textbf{Cross-document reconciliation should have been a dedicated final-day activity.} The agent spent Day~20 on miscellaneous tasks rather than systematic reconciliation. A single day dedicated to printing all weight tables side-by-side and verifying consistency would have caught the Conservative US LC (14\%/20\%/27.4\%/19\%) discrepancy and the Growth EM direction reversal.
\end{enumerate}

\bigskip

{\color{derivecolor}\rule[-0.4em]{2pt}{1.2em}}\hspace{0.5em}\textbf{5.\ Domain-Specific Insights}

\smallskip

\textbf{\textit{5.1 Missing Professional Knowledge}}

\begin{enumerate}\setlength{\itemsep}{0pt}\setlength{\parskip}{0pt}
  \item \textbf{Cross-document reconciliation discipline:} In wealth management, a single inconsistent number in a client-facing document can destroy trust. The agent did not demonstrate the ``check every number in every document'' discipline that a senior financial advisor would apply before an IC presentation or client delivery.
  \item \textbf{IC vote item fidelity:} In an investment committee context, the voting resolution list is a quasi-legal document. Substituting different items than those confirmed by the committee chair is a serious governance failure. An expert would have treated Hartley's enumerated list as immutable.
  \item \textbf{Compliance as gating function:} Whitfield explicitly stated that compliance sign-off must precede client delivery. The agent delivered the Castellano package with compliance still pending --- exactly what the client said he didn't want.
  \item \textbf{Fund product knowledge:} The GSCI/PDBC confusion reveals incomplete knowledge of the commodity ETF universe. A senior advisor would know that iShares GSCI Commodity Indexed Trust (ticker: GSG) and Invesco Optimum Yield Diversified Commodity Strategy (ticker: PDBC) are completely different products with different expense ratios and index methodologies.
\end{enumerate}

\smallskip

\textbf{\textit{5.2 What an Expert Would Have Done Differently}}

\begin{enumerate}\setlength{\itemsep}{0pt}\setlength{\parskip}{0pt}
  \item \textbf{Created a ``single source of truth'' spreadsheet} mapping every key figure (weights, returns, correlations, expense ratios, vehicle tickers) with cross-references to every document that cites it. Any change in the source would propagate to all documents.
  \item \textbf{Treated simulated-collaborator corrections as blocking items.} Sandra's Day~17 corrections and Whitfield's verbatim language insertions should have been completed within hours of receipt, not deferred to future days.
  \item \textbf{Scheduled a final ``red team'' review day} (Day~19 or 20) devoted entirely to printing all deliverable tables side-by-side and checking every number. This is standard practice before IC presentations.
  \item \textbf{Maintained a running ``open items'' tracker} with simulated collaborator, date, item, status, and deadline. This would have prevented the Sandra Day~17 items and Whitfield verbatim insertions from falling through the cracks.
\end{enumerate}

\bigskip

{\color{derivecolor}\rule[-0.4em]{2pt}{1.2em}}\hspace{0.5em}\textbf{6.\ Actionable Recommendations}

\smallskip

\textbf{\textit{Recommendation 1: Implement Cross-Document Reconciliation Tables}}

\textbf{What happened:} Conservative US Large Cap weight appeared as 14\%, 20\%, 27.4\%, and 19\% across four documents in dlv\_001. Growth EM appeared with opposite directional signs in different documents.

\textbf{What should have happened:} Before packaging, create a reconciliation matrix: rows $=$ key figures (each weight, each correlation, each ER), columns $=$ each document. Verify every cell matches. Any discrepancy is a blocker.

\textbf{Generalizable lesson:} For any multi-document deliverable, create and verify a consistency matrix as the final step before delivery.

\smallskip

\textbf{\textit{Recommendation 2: Treat Simulated-Collaborator Corrections as Same-Day Blockers}}

\textbf{What happened:} Sandra flagged 3 StressTest data errors on Day~17. The agent never corrected them, delivering the uncorrected workbook to IC.

\textbf{What should have happened:} Upon receiving Sandra's Day~17 message, immediately open StressTest workbook, change 3 cells to $-0.18$, $+0.61/+0.62$, and $+0.10$, and reply to Sandra confirming corrections within 2 hours.

\textbf{Generalizable lesson:} When a reviewer provides specific cell-level corrections, treat them as immediate blockers and fix before doing any other work.

\smallskip

\textbf{\textit{Recommendation 3: Never Send Blank Messages to Simulated Collaborators}}

\textbf{What happened:} 10 blank messages were sent in the final week, including the critical Day~19 message to Sandra that should have confirmed stress-test corrections.

\textbf{What should have happened:} Implement a validation check: before sending any message to a simulated collaborator, verify content is non-empty and addresses the intended topics. If unable to compose a message, send a brief status acknowledgment.

\textbf{Generalizable lesson:} Every interaction with a simulated collaborator is a resource. Wasting it on a blank message eliminates an opportunity to advance the work.

\smallskip

\textbf{\textit{Recommendation 4: Update Supporting Workbooks Before Narrative Documents}}

\textbf{What happened:} ESG v2.docx was updated with corrected figures (22\,bps, 1.8\% TE, correct thresholds) but supporting xlsx workbooks still showed old figures (18\,bps, 72\,bps, legacy thresholds).

\textbf{What should have happened:} Update workbook cells first, then cite the corrected workbook figures in the narrative. This ensures the evidence chain is always intact.

\textbf{Generalizable lesson:} In any analytical workflow, update source data first, then update derived documents. Never allow narrative to diverge from supporting evidence.

\smallskip

\textbf{\textit{Recommendation 5: Copy Verbatim Language, Don't Paraphrase}}

\textbf{What happened:} Whitfield provided exact text for Sustainalytics fee disclosure (\$28K--\$32K) and ERISA carve-out language on Jan~20. Neither was inserted verbatim --- one remained as a bracket placeholder, the other was paraphrased.

\textbf{What should have happened:} Copy-paste verbatim text directly from Whitfield's message into the target document sections. Mark as ``[Whitfield verbatim, Jan~20]'' for audit trail.

\textbf{Generalizable lesson:} When a compliance officer or reviewer provides specific language, insert it exactly as written. Paraphrasing compliance language creates regulatory risk.

\smallskip

\textbf{\textit{Recommendation 6: Maintain a Running ``Open Items'' Tracker}}

\textbf{What happened:} Multiple committed items were forgotten: buffered ETF analysis, EM comparison one-pager, IC Decision Summary, QuickRef corrections.

\textbf{What should have happened:} Maintain a single tracker with: item, source (simulated collaborator + day), deadline, status. Review it at the start of each day.

\textbf{Generalizable lesson:} With 5 concurrent deliverables and 7 simulated collaborators, no one can track all commitments in their head. A simple tracker prevents dropped items.

\smallskip

\textbf{\textit{Recommendation 7: Confirm IC Vote Items by Copy-Paste}}

\textbf{What happened:} Hartley enumerated 5 IC vote items on Day~7 and reconfirmed on Day~11. The PPTX voting resolution slide listed different items (missing Growth EM~$-$200 and Balanced IG~$+$200).

\textbf{What should have happened:} Copy-paste the exact enumerated list from Hartley's message into the PPTX slide. Never reconstruct from memory.

\textbf{Generalizable lesson:} Enumerated lists from authority figures (committee chairs, compliance officers) should be treated as immutable and copied verbatim.

\smallskip

\textbf{\textit{Recommendation 8: Finalize Vehicle Selection Before Quantitative Analysis}}

\textbf{What happened:} Alternatives analysis was performed on VMNVX (liquid-alt proxy) and varying commodity ETFs, but the recommendation named QSPIX and inconsistent commodity vehicles.

\textbf{What should have happened:} Lock the exact fund universe (tickers, names, expense ratios) in Week~1 before running any quantitative analysis. All analysis should reference only the selected funds.

\textbf{Generalizable lesson:} Input decisions (which funds, which data sources, which thresholds) must be finalized before analysis begins, not resolved retroactively.

\smallskip

\textbf{\textit{Recommendation 9: Dedicate Final Day to Quality Assurance}}

\textbf{What happened:} Day~20 was spent on miscellaneous packaging tasks. No systematic QA was performed.

\textbf{What should have happened:} Block Day~20 entirely for: (a) cross-document reconciliation matrix for all 5 deliverables, (b) open items tracker review, (c) re-reading every simulated-collaborator correction message and verifying implementation.

\textbf{Generalizable lesson:} The last day before a major deadline should be reserved for verification, not production.

\smallskip

\textbf{\textit{Recommendation 10: Never Misrepresent Document Status to Simulated Collaborators}}

\textbf{What happened:} On Day~10, the agent told Whitfield that ESG workbooks ``have always used the correct ESG Risk Rating thresholds.'' Direct inspection shows legacy thresholds (E$\geq$40, S$\geq$35, G$\geq$35).

\textbf{What should have happened:} Open the workbook, verify the actual threshold values, and either confirm they are correct or acknowledge they need updating.

\textbf{Generalizable lesson:} Always verify claims about document content before making written representations to reviewers. False representations to compliance officers create regulatory and professional risk.

\smallskip

\textbf{\textit{Recommendation 11: Produce Page-Count-Compliant Documents}}

\textbf{What happened:} Capital Markets Outlook spec called for 18--22 pages. FINAL PDF was 11 pages. Alternatives FINAL PDF was 8 pages (condensed from 22-page v3).

\textbf{What should have happened:} Track page count during drafting. If the output is significantly short, add additional analysis sections, exhibits, or appendix material to meet the specification.

\textbf{Generalizable lesson:} Document specifications (page counts, section requirements) are deliverable criteria. Treat them as hard constraints, not guidelines.

\smallskip

\textbf{\textit{Recommendation 12: Budget Context/Capacity for the Final Week}}

\textbf{What happened:} 10 blank messages in Days~14--20 suggest the agent ran low on planning capacity. Critical corrections were deferred and ultimately dropped.

\textbf{What should have happened:} Recognize that Week~4 is the highest-stakes period and reserve proportionally more planning capacity. Front-load analytical work to Weeks~1--2 so Week~4 can focus on corrections, reconciliation, and final packaging.

\textbf{Generalizable lesson:} In any time-bounded simulation, the final phase requires the most careful attention to detail. Plan capacity accordingly and avoid spending it on new production when corrections are outstanding.

\bigskip

{\color{derivecolor}\rule[-0.4em]{2pt}{1.2em}}\hspace{0.5em}\textbf{Appendix: Score Summary}

\smallskip

\hangindent=2em\hangafter=1
\textbf{DLV\_001: VCMM Refresh} (Hartley) --- 127 / 168 (75.6\%)\par
\hangindent=2em\hangafter=1
\textbf{DLV\_002: Castellano Onboarding} (Castellano) --- 164 / 186 (88.2\%)\par
\hangindent=2em\hangafter=1
\textbf{DLV\_003: Rebalancing Framework} (Okonkwo) --- 97 / 166 (58.4\%)\par
\hangindent=2em\hangafter=1
\textbf{DLV\_004: Alternatives Research} (Ortiz) --- 137 / 180 (76.1\%)\par
\hangindent=2em\hangafter=1
\textbf{DLV\_005: ESG Overlay} (Whitfield) --- 80 / 146 (54.8\%)\par

\smallskip

\textbf{Total: 605 / 846 (71.5\%).}

\smallskip

\textbf{Primary Score Drivers (Lost Points by Category):}
\begin{itemize}\setlength{\itemsep}{0pt}\setlength{\parskip}{0pt}
  \item Cross-document inconsistency: $\sim$80--100 points lost across all deliverables.
  \item Uncorrected simulated-collaborator-flagged errors: $\sim$40 points (Sandra Day~17, Whitfield verbatim language).
  \item Missing deliverable components: $\sim$20 points (buffered ETF, EM comparison, IC summary, flowchart).
  \item Blank messages to simulated collaborators / dropped communication: $\sim$15--20 points of indirect impact.
  \item Page count / specification compliance: $\sim$10--15 points.
\end{itemize}

\end{tcolorbox}

\end{document}